\documentclass{article}
\usepackage[12pt]{extsizes}
\usepackage[utf8]{inputenc}
\usepackage{amsmath}
\usepackage{amssymb}
\usepackage[square,sort,comma,numbers]{natbib}
\usepackage[pdftex]{graphicx}
\usepackage[dvipsnames]{xcolor}
\usepackage[hidelinks]{hyperref}
\hypersetup{
    colorlinks = true,
    linkcolor={blue!50!black},
    citecolor={blue!50!black},
    urlcolor={blue!50!black}
}
\usepackage{authblk}
\usepackage{geometry}
\usepackage{times}
\usepackage{parskip} 
\usepackage[inline]{enumitem}

\usepackage{booktabs}  
\usepackage{amsfonts}
\usepackage{mathtools}
\usepackage{url}
\usepackage{enumitem}
\usepackage{authblk}
\usepackage{comment}
\usepackage{mdframed}
\usepackage{wrapfig}
\usepackage{floatrow}
\usepackage[font=footnotesize,labelfont=bf]{caption}

\newcommand{\xhdr}[1]{\vspace{1.3mm}\noindent{{\bf #1.}}}
\newcommand{\supp}[1]{{SI Note #1}}
\newcommand{\suppf}[1]{{SI Figure #1}}

\newcommand{\hide}[1]{}
\usepackage{ntheorem}

\theoremseparator{:} 

\newmdtheoremenv[%
  backgroundcolor=white,
  linecolor=blue!60!black,
  linewidth=2pt,
  topline=true,
  rightline=false,
  skipabove=10pt,
  skipbelow=10pt,
  leftline=false]{ourexample}{Application}
  
\newmdtheoremenv[%
  backgroundcolor=gray!20,
  linecolor=red!60!black,
  linewidth=2pt,
  topline=false,
  rightline=false,
  skipabove=10pt,
  skipbelow=10pt,
  leftline=false]{ourbox}{Formulation}

\newmdtheoremenv[%
  backgroundcolor=gray!20,
  linecolor=red!60!black,
  linewidth=2pt,
  topline=false,
  rightline=false,
  skipabove=10pt,
  skipbelow=10pt,
  leftline=false]{regbox}{Box}
  
\usepackage[p,osf]{cochineal}
\usepackage[T1]{fontenc}

\usepackage[scale=.95,type1]{cabin}
\usepackage[cochineal,bigdelims,cmintegrals,vvarbb]{newtxmath}
\usepackage[zerostyle=c,scaled=.94]{newtxtt}
\usepackage[cal=boondoxo]{mathalfa}
\usepackage{microtype}

\usepackage{etoolbox}
\apptocmd{\thebibliography}{\raggedright}{}{}

\usepackage{amsmath,amsfonts,bm}

\def\eqref#1{equation~\ref{#1}}

\DeclareMathAlphabet{\mathsfit}{\encodingdefault}{\sfdefault}{m}{sl}
\SetMathAlphabet{\mathsfit}{bold}{\encodingdefault}{\sfdefault}{bx}{n}

\title{Graph Representation Learning in Biomedicine
}
\author[1,3]{Michelle M. Li}
\author[2]{Kexin Huang}
\author[3,4,5,$*$]{Marinka Zitnik}
\affil[1]{\small Bioinformatics and Integrative Genomics Program, Harvard Medical School, Boston, MA 02115, USA}
\affil[2]{Health Data Science Program, Harvard T.H. Chan School of Public Health, Boston, MA 02115, USA}
\affil[3]{Department of Biomedical Informatics, Harvard Medical School, Boston, MA 02115, USA}
\affil[4]{Broad Institute of MIT and Harvard, Cambridge, MA 02142, USA}
\affil[5]{Harvard Data Science Initiative, Cambridge, MA 02138, USA\vspace{2mm}}
\affil[$*$]{Correspondence: \href{mailto:marinka@hms.harvard.edu}{marinka@hms.harvard.edu}}
\date{}

\begin{document}

\maketitle

\begin{abstract}
\noindent 
Biomedical networks (or graphs) are universal descriptors for systems of interacting elements, from molecular interactions and disease co-morbidity to healthcare systems and scientific knowledge. Advances in artificial intelligence, specifically deep learning, have enabled us to model, analyze, and learn with such networked data. In this review, we put forward an observation that long-standing principles of systems biology and medicine---while often unspoken in machine learning research---provide the conceptual grounding for representation learning on graphs, explain its current successes and limitations, and even inform future advancements. We synthesize a spectrum of algorithmic approaches that, at their core, leverage graph topology to embed networks into compact vector spaces. We also capture the breadth of ways in which representation learning has dramatically improved the state-of-the-art in biomedical machine learning. Exemplary domains covered include identifying variants underlying complex traits, disentangling behaviors of single cells and their effects on health, assisting in diagnosis and treatment of patients, and developing safe and effective medicines.

\end{abstract}

\section{Introduction}
Networks (or graphs) are pervasive in biology and medicine, from molecular interaction maps to population-scale social and health interactions. With the multitude of bioentities and associations that can be described by networks, they are prevailing representations of biological organization and biomedical knowledge. For instance, edges in a regulatory network can indicate causal activating and inhibitory relationships between genes~\cite{qiu2020inferring}; edges between genes and diseases can indicate genes that are `upregulated by', `downregulated by', or `associated with' a disease \cite{nicholson2020constructing}; and edges in a knowledge network built from electronic health records (EHR) can indicate co-occurrences of medical codes across patients~\cite{robinson2008human, schriml2012disease,hong2021clinical}. The ability to model all biomedical discoveries to date---even overlay patient-specific information---in a unified data representation has driven the development of artificial intelligence, specifically deep learning, for networks. In fact, the diversity and multimodality in networks not only boost performance of predictive models, but importantly enable broad generalization to settings not seen during training~\cite{gysi2021network} and improve model interpretability~\cite{nelson2019integrating, chen2020pathomic}. Nevertheless, interactions in networks give rise to a bewildering degree of complexity that can likely only be fully understood through a holistic and integrated view~\cite{callahan2020knowledge, doi:10.1056/NEJMe078114, mungall2017monarch}. As a result, systems biology and medicine---upon which deep learning on graphs is founded---have identified over the last two decades {\em organizing principles} that govern networks~\cite{goh2007human, barabasi2011network, hu2016network, zitnik2019evolution}.

\begin{figure}[t]
\centering
\includegraphics[width=\textwidth]{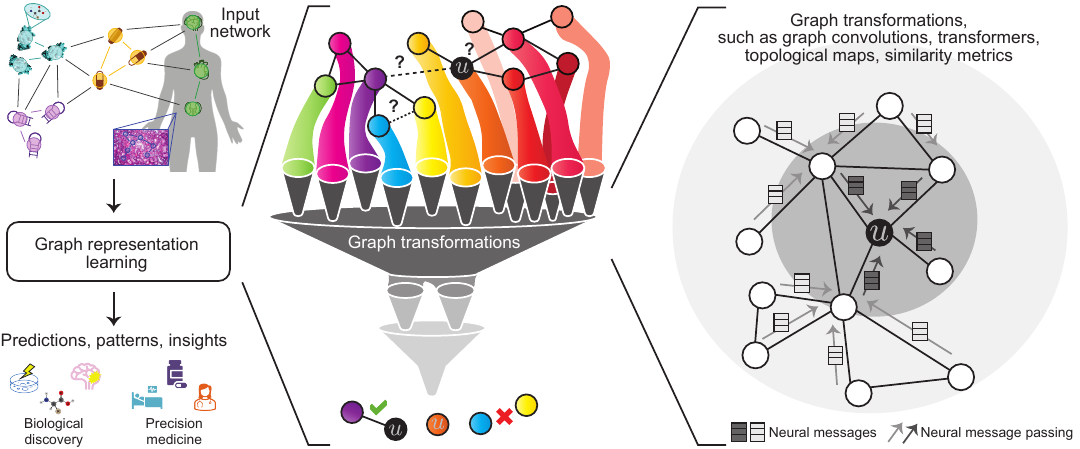}
\caption{\textbf{Representation learning for networks in biology and medicine.} Given a biomedical network, a representation learning method transforms the graph to extract patterns and leverage them to produce compact vector representations that can be optimized for the downstream task. The far right panel shows a local 2-hop neighborhood around node $u$, illustrating how information (e.g., neural messages) can be propagated along edges in the neighborhood, transformed, and finally aggregated at node $u$ to arrive at the $u$'s embedding.
}
\label{fig:GML-overview}
\end{figure}

The organizing principles governing networks {\em link network structure to molecular phenotypes, biological roles, disease, and health}, thus providing the conceptual grounding that, we argue, can explain the successes (and limitations) of graph representation learning and inform future development of the field. Here, we exemplify how a series of such principles has uncovered disease mechanisms. First, interacting entities are typically more similar than non-interacting entities, as implicated by the local hypothesis~\cite{barabasi2011network}. In protein interaction networks, for instance, mutations in interacting proteins often lead to similar diseases~\cite{barabasi2011network}. Given by the shared components and disease module hypotheses~\cite{barabasi2011network}, cellular components associated with the same phenotype tend to cluster in the same network neighborhood~\cite{agrawal2018}. Further, essential genes are often found in hubs of a molecular network whereas non-essential genes (e.g., those associated with disease) are located on the periphery~\cite{barabasi2011network}. Thus, the network parsimony principle dictates that shortest molecular paths between known disease-associated components tend to correlate with causal molecular pathways \cite{barabasi2011network}. To this day, these hypotheses and principles continue to drive discoveries.

We posit that {\em representation learning can realize network biomedicine principles}. Its core idea is to learn how to represent nodes (or larger graph structures) in a network as points in a low-dimensional space, where the geometry of this space is optimized to reflect the structure of interactions between nodes. More concretely, representation learning specifies deep, non-linear transformation functions that map nodes to points in a compact vector space, termed {\em embeddings}. Such functions are optimized to embed the input network so that nodes with similar network neighborhoods are embedded close together in the embedding space, and algebraic operations performed in this learned space reflect the network's topology. To provide concrete connections between graph representation learning and systems biology and medicine: nodes in the same positional regions should have similar embeddings due to the local hypothesis (e.g.,~highly similar pairs of protein embeddings suggest similar phenotypic consequence); node embeddings can capture whether the nodes lie within a hub based on their degree, an important aspect of local neighborhood (e.g.,~strongly clustered gene embeddings indicate essential housekeeping roles); and given by the shared components hypothesis, two nodes with significantly overlapping sets of network neighbors should have similar embeddings due to shared message passing (e.g.,~highly similar disease embeddings imply shared disease-associated cellular components). Hence, artificial intelligence methods that produce representations can be thought of as differentiable engines of key network biomedicine principles.

Our survey provides an exposition of graph artificial intelligence capability and highlights important applications for deep learning on biomedical networks. Given the prominence of graph representation learning, specific aspects of it have been covered extensively. However, existing reviews independently discuss deep learning on structured data~\cite{camacho2018next, zhang2020deep}; graph neural networks~\cite{hamilton2017representation, hamilton2020graph,  wu2020comprehensive}; representation learning for homogeneous and heterogeneous graphs~\cite{chen2020graph, li2020network, yue2020graph}, solely heterogeneous graphs~\cite{dong2020heterogeneous}, and dynamic graphs~\cite{kazemi2020representation}; data fusion~\cite{zitnik2019machine}; network propagation~\cite{cowen2017network}; topological data analysis~\cite{blevins2020topology}; and creation of biomedical networks~\cite{blevins2020topology, callahan2020knowledge, koutrouli2020guide, liu2020computational, rai2018network}. Biomedically-focused reviews survey graph neural networks exclusively on molecular generation~\cite{david2020molecular, wieder2020compact}, single-cell biology~\cite{hetzel2021graph}, drug discovery and repurposing~\cite{jimenez2020drug, sun2020graph, gaudelet2021utilizing, maclean2021knowledge, zeng2022toward}, or histopathology~\cite{ahmedt2021survey}. Other reviews tend to focus solely on graph neural networks, excluding other graph representation learning approaches or do not consider patient-centric methods~\cite{muzio2021biological}. In contrast, our survey unifies graph representation learning approaches across molecular, genomic, therapeutic, and precision medicine areas.

\section{Graph representation learning}\label{sec:graphrep}

Graph theoretic techniques have fueled discoveries, from uncovering relationships between diseases~\cite{guo2019analysis, le2016ontology, menche2015uncovering, sumathipala2019network} to repurposing drugs~\cite{cheng2019network, cheng2019genome, gysi2021network}. Further algorithmic innovations, such as random walks~\cite{chen2020prediction, wong2020mipdh, yang2018hergepred}, kernels~\cite{geng2020iscore}, and network propagation~\cite{veselkov2019hyperfoods}, have also played a role in capturing structural information in networks. Feature engineering, the process of extracting predetermined features from a network to suit a user-specified machine learning method~\cite{zheng2018feature}, is a common approach for machine learning on networks, including but not limited to hard-coding network features (e.g., higher-order structures, network motifs, degree counts, and common neighbor statistics) and feeding the engineered feature vectors into a machine learning model. While powerful, it can be challenging to hand engineer optimally-predictive features across diverse types of networks and applications~\cite{zhang2020deep}.

For these reasons, {\em graph representation learning}, the idea of automatically learning optimal features for networks, has emerged as a leading artificial intelligence approach for networks. Graph representation learning is challenging because graphs contain complex topographical structure, have no fixed node ordering or reference points, and are comprised of many different kinds of entities (nodes) and various types of interactions (edges) relating them to each other. Classic deep learning methods are unable to consider such diverse structural properties and rich interactions, which are the essence of biomedical networks, because classic deep methods are designed for fixed-size grids (e.g., images and tabular datasets) or optimized for text and sequences. Akin to how deep learning on images and sequences has revolutionized image analysis and natural language processing, graph representation learning is poised to transform the study of complex systems.

\begin{figure}
    \centering
    \includegraphics[width=0.9\textwidth]{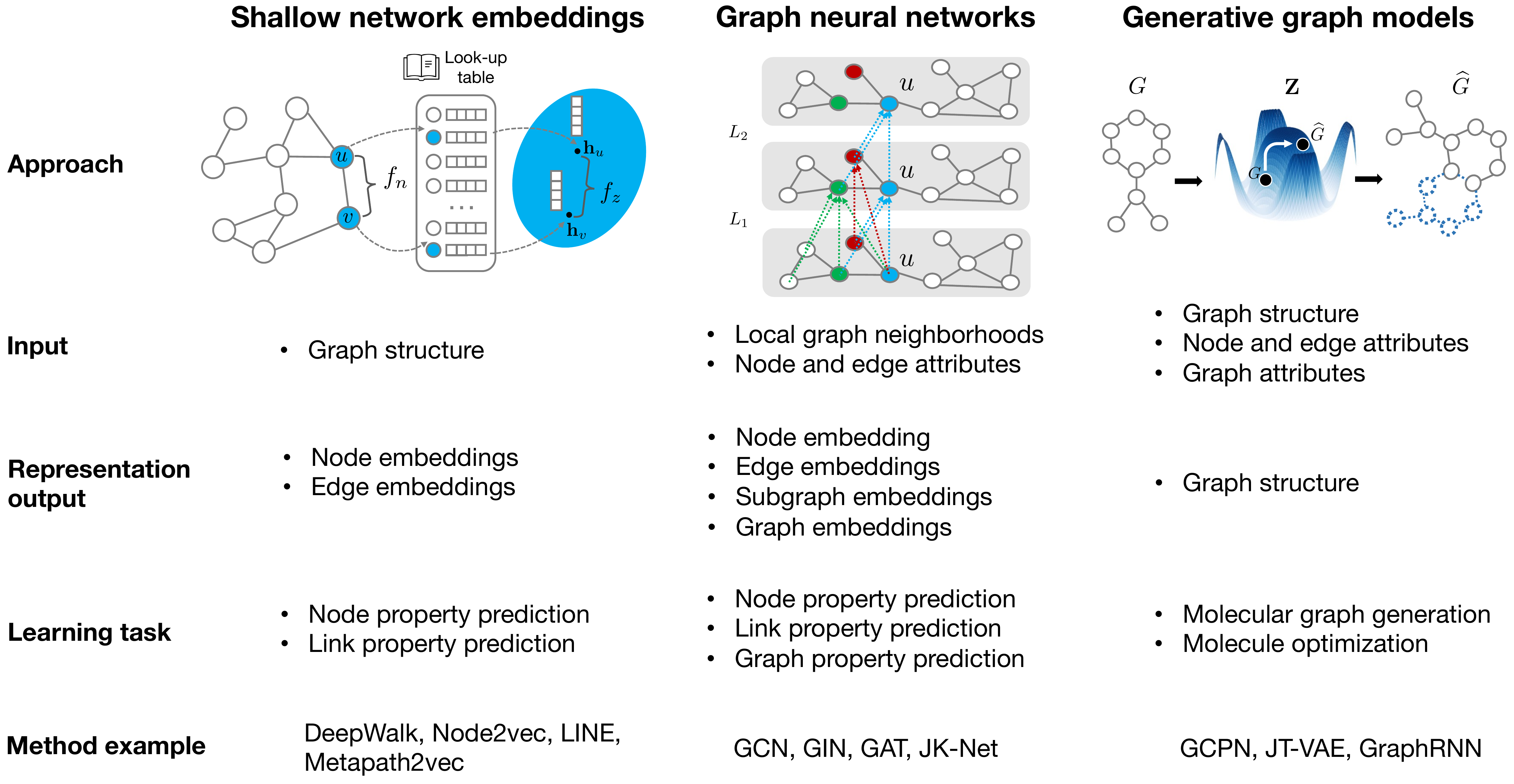}
    \caption{\textbf{Predominant paradigms in graph representation learning.} 
    \textbf{(a)} Shallow network embedding methods generate a dictionary of representations $\mathbf{h}_u$ for every node $u$ that preserves the input graph structure information. This is achieved by learning a mapping function $f_z$ that maps nodes into an embedding space such that nodes with similar graph neighborhoods measured by function $f_n$ get embedded closer together (Section~\ref{sec:shallow}). 
    Given the learned embeddings, an independent decoder method can optimize embeddings for downstream tasks, such as node or link property prediction. Method examples include DeepWalk~\cite{perozzi2014deepwalk}, Node2vec~\cite{node2vec}, LINE~\cite{tang2015line}, and Metapath2vec~\cite{metapath2vec}.
    \textbf{(b)} In contrast with shallow network embedding methods, graph neural networks can generate representations for any graph element by capturing both network structure and node attributes and metadata. The embeddings are generated through a series of non-linear transformations, i.e., message-passing layers ($L_k$ denotes transformations at layer $k$), that iteratively aggregate information from neighboring nodes at the target node $u$. GNN models can be optimized for performance on a variety of downstream tasks (Section~\ref{sec:gnn}). Method examples include GCN~\cite{gcn}, GIN~\cite{gin}, GAT~\cite{gat}, and JK-Net~\cite{xu2018representation}.
    \textbf{(c)} Generative graph models estimate a distribution landscape $\mathbf{Z}$ to characterize a collection of distinct input graphs. They use the optimized distribution to generate novel graphs $\widehat{G}$ that are predicted to have desirable properties, e.g., a generated graph can be represent a molecular graph of a drug candidate. Generative graph models use graph neural networks as encoders and produce graph representations that capture both network structure and attributes (Section~\ref{sec:generative}). Method examples include GCPN~\cite{gcpn}, JT-VAE~\cite{jtvae}, and GraphRNN~\cite{you2018graphrnn}.
    \suppf{1} and SI Note 3 outline other representation learning techniques.
    }
    \label{fig:method}
\end{figure}

Graph representation learning methods generate vector representations for graph elements such that the learned  representations, i.e., {\em embeddings,} capture the structure and semantics of networks, along with any downstream supervised task, if any (Box~\ref{box:notation}). Graph representation learning encompasses a wide range of methods, including manifold learning, topological data analysis, graph neural networks and generative graph models (Figure~\ref{fig:method}). We next describe graph elements and outline main artificial intelligence tasks on graphs (Box \ref{box:notation}). We then outline graph representation learning methods (Section \ref{sec:shallow}-\ref{sec:generative}).

\clearpage

\begin{regbox}
\normalfont\textit{\textbf{Fundamentals of graph representation learning}} \label{box:notation}

\xhdr{Elements of graphs}
Graph $G = (\mathcal{V}, \mathcal{E})$ consists of nodes $v \in \mathcal{V}$ and edges or relations $e^r_{u,v} \in \mathcal{E}$ connecting nodes $u$ and $v$ via a relationship of type $r$. 
Subgraph $S = (\mathcal{V}_S, \mathcal{E}_S)$ is a subset of a graph $G$, where $\mathcal{V}_S \subseteq \mathcal{V}$ and $\mathcal{E}_S \subseteq \mathcal{E}$. Adjacency matrix $\mathbf{A}$ is used to represent a graph, where each entry $\mathbf{A}_{u,v}$ is 1 if nodes $u, v$ are connected, and 0 otherwise. $\mathbf{A}_{u,v}$ can also be the edge weight between nodes $u,v$. Homogeneous graph is a graph with a single node and edge type. In contrast, heterogeneous graph consists of nodes of different types (node type set $\mathcal{A}$) connected by diverse kinds of edges (edge type set $\mathcal{R}$). Node attribute vector $\mathbf{x}_u \in \mathbb{R}^d$ describes side information and metadata of node $u$. The node attribute matrix $\mathbf{X} \in \mathbb{R}^{n \times d}$ brings together attribute vectors for all nodes in the graph. Similarly, edge attributes $\mathbf{x}_{u,v}^e \in \mathbb{R}^c$ for edge $e_{u,v}$ can be taken together to form an edge attribute matrix $\mathbf{X}^e \in \mathbb{R}^{m \times c}$. A path from node $u_1$ to node $u_k$ is given by a sequence of edges $u_1 \xrightarrow{e_{1,2}} u_2 \cdots u_{k-1} \xrightarrow{e_{k-1,k}} u_k$. For node $u$, we denote its neighborhood $\mathcal{N}(u)$ as nodes directly connected to $u$ in $G$, and the node degree is the size of $\mathcal{N}(u)$. The $k$-hop neighborhood of node $u$ is the set of nodes that are exactly $k$ hops away from node $u$, that is, $\mathcal{N}^k(u) = \{v | d(u, v) = k\}$ where $d$ denotes the shortest path distance (\supp{1}).

\xhdr{Artificial intelligence tasks on graphs}
To extract this information from networks, classic machine learning approaches rely on summary statistics (e.g., degrees or clustering coefficients) or carefully engineered features to measure network structures (e.g., network motifs). In contrast, representation learning approaches automatically learn to encode networks into low-dimensional representations (i.e., embeddings) using transformation techniques based on deep learning and nonlinear dimensionality reduction. The flexibility of learned representations shows in a myriad of tasks that representations can be used for (\supp{2}):

\begin{itemize}[leftmargin=*]\itemsep0em 
    \item \textbf{Node, link, and graph property prediction:} The objective is to learn representations of  graph elements, such as nodes, edges, subgraphs, and entire graphs. Representations are optimized so that performing algebraic operations in the embedding space reflects the graph's topology. Optimized representations can be fed into models to predict properties of graph elements, such as the function of proteins in an interactome network (i.e., node classification task), the binding affinity of a chemical compound to a target protein (i.e., link prediction task), and the toxicity profile of a candidate drug (i.e., graph classification task).
    \item \textbf{Latent graph learning:} Graph representation learning exploits relational inductive biases for data that come in the form of  graphs. In some settings, however, the graphs are not readily available for learning. This is typical for many biological problems, where graphs such as gene regulatory networks are only partially known. Latent graph learning is concerned with inferring the graph from the data. The latent graph can be application-specific and optimized for the downstream task. Further, such a graph might be as important as the task itself, as it can convey insights about the data and offer a way to interpret the results.
    \item \textbf{Graph generation:} The objective is to generate a graph $G$ representing a biomedical entity that is likely to have a property of interest, such as high druglikeness. The model is given a set of graphs $\mathcal{G}$ with such a property and is tasked with learning a non-linear mapping function characterizing the distribution of graphs in $\mathcal{G}$. The learned distribution is used to optimize a new graph $G$ with the same property as input graphs.

\end{itemize}

\end{regbox}

\subsection{Shallow graph embedding approaches} \label{sec:shallow} 

Shallow embedding methods optimize a compact vector space such that points that are close in the graph are mapped to nearby points in the embedding space, which is measured by a predefined distance function or an outer product. These approaches are transductive embedding methods where the encoder function is a simple embedding lookup (Figure~\ref{fig:method}). More concretely, the methods have three steps: (1) \textit{Mapping to an embedding space.} Given a pair of nodes $u$ and $v$ in graph $G$, we specify an encoder, a learnable function $f$ that maps nodes to embeddings $\mathbf{h}_u$ and $\mathbf{h}_v$. (2) \textit{Defining graph similarity.} We define the graph similarity $f_n(u,v)$, for example, measured by distance between $u$ and $v$ in the graph, and the embedding similarity $f_z(\mathbf{h}_u,\mathbf{h}_v)$, for example, an Euclidean distance function or pairwise dot-product. (3) \textit{Computing loss.} Then, we define the loss $\mathcal{L}(f_n(u,v), f_z(\mathbf{h}_u,\mathbf{h}_v))$, which quantifies how the resulting embeddings preserve the desired input graph similarity. Finally, we apply an optimization procedure to minimize the loss $\mathcal{L}(f_n(u,v), f_z(\mathbf{h}_u,\mathbf{h}_v))$. The resulting encoder $f$ is a graph embedding method that serves as a shallow embedding lookup and considers the graph structure only in the loss function.

Shallow embedding methods vary given various definitions of similarities. For example, shortest path length between nodes is often used as the network similarity and dot product as the embedding similarity. Perozzi {\em et al.}~\cite{perozzi2014deepwalk} define similarity as co-occurrence in a series of random walks of length $k$. Unsupervised techniques that predict which node belongs to the walk, such as skip-gram~\cite{mikolov2013distributed}, are then applied on the walks to generate embeddings. Grover {\em et al.}~\cite{node2vec} propose an alternative way for walks on graphs, using a combination of depth-first search and breadth-first search. In heterogeneous graphs, information on the semantic meaning of edges, i.e., relation types, can be important. Knowledge graph methods expand similarity measures to consider relation types~\cite{transe,rescal,trouillon2016complex,sun2019rotate,distmult,metapath2vec}. When shallow embedding models are trained, the resulting embeddings can be fed into separate models to be optimized towards a specific classification or regression task.

\subsection{Graph neural networks} \label{sec:gnn} 

Graph neural networks (GNNs) are a class of neural networks designed for graph-structured datasets (Figure~\ref{fig:method}). They learn compact representations of graph elements, their attributes, and supervised labels if any. A typical GNN consists of a series of propagation layers~\cite{enn-s2s}, where layer $l$ carries out three operations: (1) \textit{Passing neural messages.} The GNN computes a message $\mathbf{m}^{(l)}_{u,v} = \textsc{Msg}(\mathbf{h}_u^{(l-1)}, \mathbf{h}_v^{(l-1)})$ for every linked nodes $u,v$ based on their embeddings from the previous layer $\mathbf{h}_u^{(l-1)}$ and $\mathbf{h}_v^{(l-1)}$. (2) \textit{Aggregating neighborhoods.} The messages between node $u$ and its neighbors $\mathcal{N}_u$ are aggregated as $\widehat{\mathbf{m}}^{(l)}_{u} = \textsc{Agg}({\mathbf{m}^{(l)}_{uv} | v \in \mathcal{N}_u})$. (3) \textit{Updating representations.} Non-linear transformation is applied to update node embeddings as $\mathbf{h}^{(l)}_u = \textsc{Upd}(\widehat{\mathbf{m}}^{(l)}_{u}, \mathbf{h}^{(l-1)}_u)$ using the aggregated message and the embedding from the previous layer. In contrast to shallow embeddings, GNNs can capture higher-order and non-linear patterns through multi-hop propagation within several layers of neural message passing. Additionally, GNNs can optimize supervised signals and graph structure simultaneously, whereas shallow embedding approaches require a two-stage approach to achieve the same.

A myriad of GNN architectures define different messages, aggregation, and update schemes to derive deep graph embeddings~\cite{NeuralFP,set2set,defferrard2016convolutional,gcn}. For example, \cite{gat,hu2020heterogeneous,yun2019graph,yan2018spatial,choi2020learning} assign importance scores for nodes during neighborhood aggregation such that more important nodes play a larger effect in the embeddings. \cite{JK-Net,abu2019mixhop} improve GNNs' ability to capture graph structural information by posing structural priors, such as a higher-order adjacency matrix. Graph pooling techniques~\cite{diffpool} learn abstract topological structures. GNNs designed for molecules~\cite{schnet,dimenet} inject physics-based scores and domain knowledge into propagation layers.

As biomedical networks can be multimodal and massive, special consideration is needed to scale GNNs to large and heterogeneous networks. To this end, \cite{chiang2019cluster,GraphSAINT} developed sampling strategies to intelligently select small subsets of the whole local network neighborhoods and use them in training GNN models. To tackle heterogeneous relations, \cite{RGCN,HAN,hu2020heterogeneous} designed aggregation transformations to fuse diverse types of relations and attributes. Recent architectures describe dynamic message passing~\cite{pareja2020evolvegcn,rossi2020temporal,hu2020heterogeneous} to deal with evolving and time-varying graphs and few-shot learning~\cite{huang2020graph} or self-supervised strategies~\cite{Pretraining,you2020does} to deal with graphs that are poorly annotated and have limited label information.

\subsection{Generative graph models} \label{sec:generative} 

Generative graph models generate new node and edge structures---even entire graphs---that are likely to have desired properties, such as novel molecules with acceptable toxicity profiles (Figure~\ref{fig:method}). Traditionally, network science models can generate graphs using deterministic or probabilistic rules. For instance, the Erd\"{o}s-R\'{e}nyi model~\cite{erdHos1960evolution} keeps iteratively adding random edges according to a predefined probability starting from an empty graph; the Barab\'{a}si-Albert model~\cite{albert2002statistical} grows a graph by adding nodes and edges such that the resulting graph has a power-law degree distribution often observed in real-world networks; the configuration model~\cite{barabasi2016network} adds edges based on a predefined node degree sequences to generate graphs with arbitrary degree distributions. While powerful as random graph generators, such models cannot optimize graph structure based on properties of interest. 

Deep generative models address the challenge by estimating distributional graph properties based on a dataset of graphs $\mathcal{G}$ and inferring graph structures using such optimized distributions. In particular, a generation graph model first learns a latent distribution $P(Z|\mathcal{G})$ that characterizes the input graph set $\mathcal{G}$. Then, conditioned on this distribution, it decodes a new graph, i.e., generates a new graph $\widehat{G}$. There are different ways to encode the input graphs and learn the latent distribution, such as through variational autoencoders~\cite{VGAE,gomez2018automatic,jtvae} and generative adversarial networks~\cite{wang2017graphgan}. Decoding to generate a novel graph presents a unique challenge compared to an image or text since a graph is discrete, is unbounded in structure and size, and has no particular ordering of nodes. Common practices to generate new graphs include (1) predicting a probabilistic fully-connected graph and then using graph matching to find the optimal subgraph~\cite{simonovsky2018graphvae}; (2) decomposing a graph into a tree of subgraphs structure and generating a tree structure instead, followed by assembles of subgraphs)~\cite{jtvae}; and (3) sequentially sampling new nodes and edges~\cite{you2018graphrnn,gcpn}.

\section{Application areas in biology and medicine}\label{sec:areas}

\begin{figure}
    \centering
    \includegraphics[width = \textwidth]{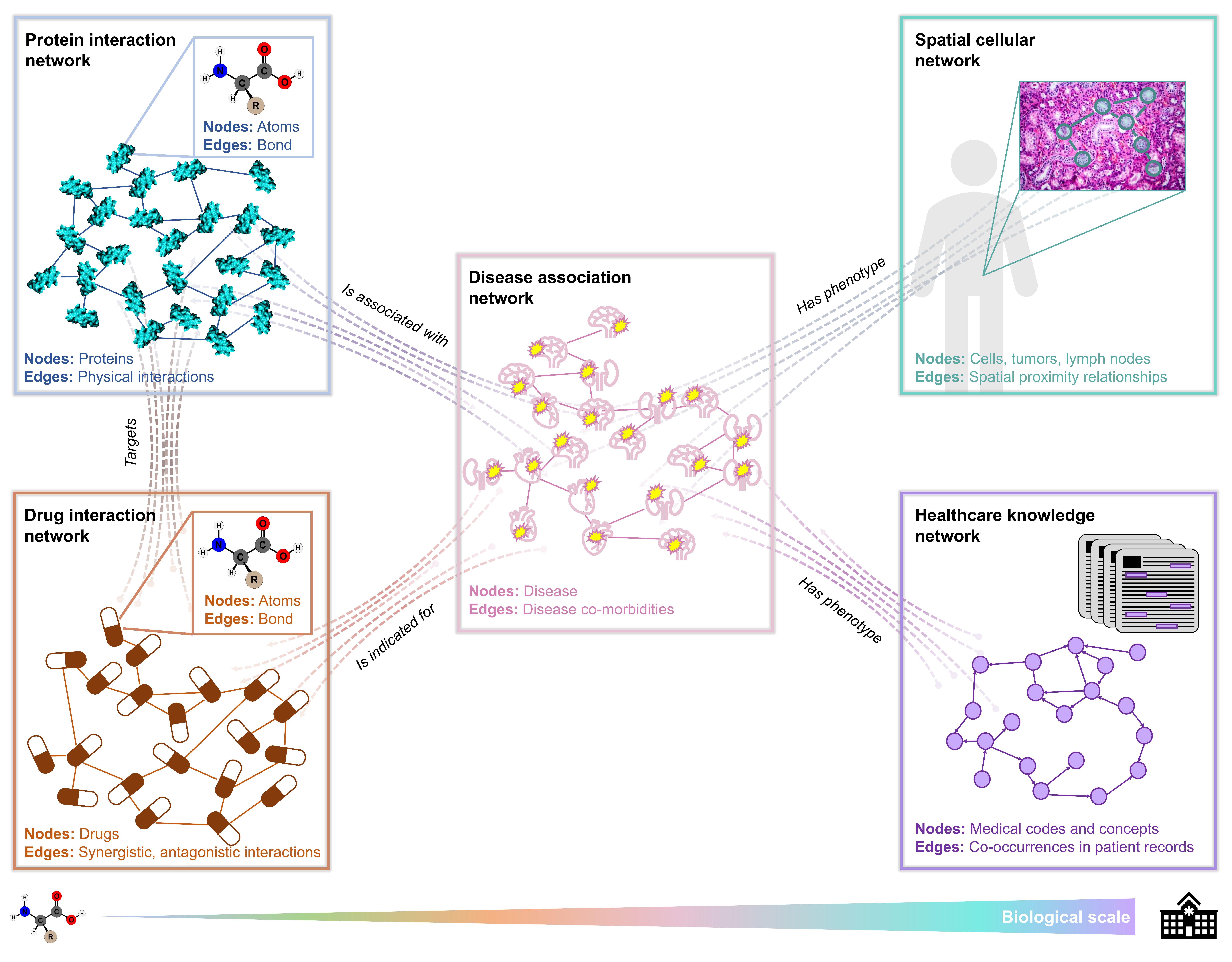}
    \caption{\textbf{Overview of biomedical applications areas.} Networks are prevalent across biomedical areas, from the molecular level to the healthcare systems level. Protein structures and therapeutic compounds can be modeled as a network where nodes represent atoms and edges indicate a bond between pairs of atoms. Protein interaction networks contain nodes that represent proteins and edges that indicate physical interactions (top left). Drug interaction networks are comprised of drug nodes connected by synergistic or antagonistic relationships (bottom left). Protein- and drug-interaction networks can be combined using an edge type that signifies a protein being a ``target" of a drug (left). Disease association networks often contain disease nodes with edges representing co-morbidity (middle). Edges exist between proteins and diseases to indicate proteins (or genes) associated with a disease (top middle). Edges exist between drugs and diseases to signify drugs that are indicated for a disease (bottom middle). Patient-specific data, such as medical images (e.g., spatial networks of cells, tumors, and lymph nodes) and EHRs (e.g., networks of medical codes and concepts generated by co-occurrences in patients' records), are often integrated into a cross-domain knowledge graph of proteins, drugs, and diseases (right). With such vast and diverse biomedical networks, we can derive fundamental insights about biology and medicine while enabling personalized representations of patients for precision medicine. Note that there are many other types of edge relations; ``targets," ``is associated with," ``is indicated for," and ``has phenotype" are a few examples.}

    \label{fig:applications}
\end{figure}

Biomedical data involve rich multimodal and heterogeneous interactions that span from molecular to societal levels (Figure \ref{fig:applications}). Unlike machine learning approaches designed to analyze data modalities like medical images and biological sequences, graph representation learning methods are uniquely able to leverage structural information in multimodal datasets~\cite{yang2020graph}. 

Starting at the \textbf{molecular level (Section~\ref{sec:proteins})}, molecular structure is translated from atoms and bonds into nodes and edges, respectively. Physical interactions or functional relationships between proteins also naturally form a network. Given by the key organizing principles that govern network medicine---for instance, the local hypothesis and the shared components hypothesis~\cite{barabasi2011network}---whether an unknown protein clusters in a particular neighborhood of and shares direct neighbors with known proteins are informative of its binding affinity, function, etc~\cite{huang2020skipgnn}. Grounded in network medicine principles, graph machine learning is commonly applied to learn molecular representations of proteins and their physical interactions for predicting protein function. 

At the \textbf{genomic level (Section~\ref{sec:genomics})}, genetic elements are incorporated into networks by extracting coding genes' co-expression information from transcriptomic data. Single-cell and spatial molecular profiling have further enabled the mapping of genetic interactions at the cellular and tissue level. Investigating the cellular circuitry of molecular functions in the resulting gene co-expression or regulatory networks can uncover disease mechanisms. For instance, as implicated by the network parsimony principle~\cite{barabasi2011network}, shortest path length in a molecular network between disease-associated components often correlates with causal molecular pathways~\cite{menche2015uncovering, agrawal2018}. Learned embeddings---using graph representation learning methods---that capture genome-wide interactions has enhanced disease predictions, even at the resolution of tissues and single cells. 

At the \textbf{therapeutics level (Section~\ref{sec:therapeutics})}, networks are composed of drugs (e.g.,~small compounds), proteins, and diseases to allow the modeling of drug-drug interactions, binding of drugs to target proteins, and identification of drug repurposing opportunities. For example, by the corollary of the local hypothesis~\cite{barabasi2011network}, the topology of drug combinations are indicative of synergistic or antagonistic relationships~\cite{yin2014synergistic}. Learning the topology of graphs containing drug, protein, and disease nodes has improved predictions of candidate drugs for treating a disease, identification of potential off-target effects, and prioritization of novel drug combinations. 

Finally, at the \textbf{healthcare level (Section~\ref{sec:healthcare})}, patient records, such as medical images and EHRs, can be represented as networks and incorporated into protein, disease, and drug networks. As an example, given by three network medicine principles, the local hypothesis, the shared components hypothesis, and the disease module hypothesis~\cite{barabasi2011network}, rare disease patients with common neighbors and topology likely have similar phenotypes and even disease mechanisms~\cite{alsentzer2020subgraph, buphamalai2021network}. Graph representation learning methods have been successful in integrating patient records with molecular, genomic, and disease networks to personalize patient predictions.

\section{Graph representation learning for molecules} \label{sec:proteins}

Graph representation learning has been widely used for predicting protein interactions and function~\cite{fan2020pseudo2go, kearnes2016molecular}. Specifically, the inductive ability of graph convolution neural networks to generalize to data points unseen during model training (Section~\ref{sec:gnn}) and even generate new data points from scratch graph by decoding latent representation from the embedding space (Section~\ref{sec:generative}) has enabled the discovery of new molecules, interactions, and functions~\cite{dutil2018towards, GraphSAGE, yang2020graph}.

\subsection{Modeling protein molecular graphs} \label{sec:prot_struc}

Computationally elucidating protein structure has been an ongoing challenge for decades~\cite{david2020molecular}. Since protein structures are folded into complex 3D structures, it is natural to represent them as graphs. For example, we construct a contact distance graph where nodes are individual residues and edges are determined by a physical distance threshold~\cite{huan2005comparing}. Edges can also be defined by the ordering of amino acids in the primary sequence~\cite{huan2005comparing}. Spatial relationships between residues (e.g.,~distances, angles) may be used as features for edges~\cite{fout2017protein}.

We then model the 3D protein structures by capturing dependencies in their sequences of amino acids (e.g.,~applying GNNs (Section \ref{sec:gnn}) to learn each node's local neighborhood structure) in order to generate protein embeddings~\cite{protein_design, fout2017protein}. After learning short- and long-range dependencies across sequences corresponding to their 3D structures to produce embeddings of proteins, we can predict primary sequences from 3D structures~\cite{protein_design}. Alternatively, we can use a hierarchical process of learning atom connectivity motifs to capture molecular structure at varying levels of granularity (e.g.,~at the motif-, connectivity-, and atomic-levels) in our protein embeddings, with which we can generate new 3D structures---a difficult task due to computational constraints of being both generalizable across different classes of molecules and flexible to a wide range of sizes~\cite{jin2020hierarchical}.
As the field of machine learning for molecules is vast, we refer readers to existing reviews on molecular design~\cite{elton2019deep, david2020molecular}, graph generation~\cite{guo2020systematic}, and molecular properties prediction~\cite{david2020molecular, wieder2020compact}, and Section~\ref{sec:mol_struc} on therapeutic compound design and generation.

\subsection{Quantifying protein interactions} \label{sec:prot_interact} 

Many have integrated various data modalities, including chemical structure, binding affinities, physical and chemical principles, and amino acid sequences, to improve protein interaction quantification~\cite{david2020molecular}. GNNs (as described in Section \ref{sec:prot_struc}) are commonly used to generate representations of proteins based on chemical (e.g.,~free electrons' and protons donors' locations) and geometric (e.g.,~distance-dependent curvature) features to predict protein pocket-ligand and protein-protein interactions sites~\cite{masif}; intra- and inter-molecule residue contact graphs to predict intra- and inter-molecular energies, binding affinities, and quality measures for a pair of molecular complexes \cite{cao2020energy}; and ligand- and receptor-protein graphs to predict whether a pair of residues from the ligand and receptor proteins is a part of an interface~\cite{fout2017protein}. Combining evolutionary, topological, and energetic information about molecules enables the scoring of docked conformations based on the similarity of random walks simulated on a pair of protein graphs (refer to graph kernel metrics in \supp{3})~\cite{geng2020iscore}.

Due to experimental and resource constraints, the most updated PPI networks are still limited in their number of nodes (proteins) and edges (physical interactions)~\cite{luck2020reference}. Topology-based methods have been shown to capture and leverage the dynamics of biological systems to enrich existing PPI networks~\cite{li2019towards}. Predominant methods first apply graph convolutions (Section \ref{sec:gnn}) to aggregate structural information in the graphs of interest (e.g.,~PPI networks, ligand-receptor networks), then use sequence modeling to learn the dependencies in amino acid sequences, and finally concatenate the two outputs for predicting the presence of physical interactions~\cite{liu2020deep, yang2020graph}. Interestingly, such concatenated outputs have been treated as ``image" inputs to CNNs~\cite{liu2020deep}, demonstrating the synergy of graph- and non-graph based machine learning methods. Similar graph convolution methods are also used to remove less credible interactions, thereby constructing a more reliable PPI network~\cite{yao2020denoising}.

\subsection{Interpreting protein functions and cellular phenotypes} \label{sec:prot_func} 

Characterizing a protein's function in specific biological contexts is a challenging and experimentally intensive task~\cite{moreau2012computational, vzitnik2015gene}. However, innovating graph representation learning techniques to represent protein structures and interactions has facilitated protein function prediction~\cite{zhou2021functions}, especially when we leverage existing gene ontologies and transcriptomic data.

Gene Ontology (GO) terms~\cite{gene2019gene} are a standardized vocabulary for describing molecular functions, biological processes, and cellular locations of gene products~\cite{zhou2019deepgoa}. They have been built as a hierarchical graph that GNNs then leverage to learn dependencies of the terms~\cite{zhou2019deepgoa}, or directly used as protein function labels~\cite{fan2020graph2go, fan2020pseudo2go}. In the latter case, we typically construct sequence similarity networks, combine them with PPI networks, and integrate protein features (e.g.~amino acid sequence, protein domains, subcellular location, gene expression profiles) to predict protein function~\cite{fan2020graph2go, fan2020pseudo2go}. Others have even created gene interaction networks using transcriptomic data~\cite{dutil2018towards, hasibi2020predicting} to capture context-specific interactions between genes, which PPI networks lack.

Alternative graph representation learning methods for predicting protein function include defining  diffusion-based distance metrics on PPI networks for predicting protein function~\cite{cao2013going}; using the theory of topological persistence to compute signatures of a protein based on its 3D structure~\cite{dey2018protein}; and applying topological data analysis to extract features from protein contact networks created from 3D coordinates~\cite{martino2018supervised} (\supp{3}). Many have also adopted an attention mechanism for protein sequence embeddings generated by a Bidirectional Encoder Representations from Transformers (BERT) model to enable interpretability~\cite{nambiar2020transforming, vig2020bertology}, showcasing the synergy of graph-based and language models.

\clearpage

\begin{regbox}
\normalfont\textit{\textbf{Learning multi-scale representations of proteins and cell types (Figure~\ref{fig:case-studies}a)}}
\label{sec:mol_app} 

\xhdr{Graph dataset}
Activation of gene products can vary considerably across cells. Single-cell transcriptomic and proteomic data captures the heterogeneity of gene expression across diverse types of cells~\cite{stoeckius2017simultaneous, kondratova2019multiscale}. With the help of GNNs, we inject cell type specific expression information into our construction of cell type specific gene interaction networks~\cite{rizvi2017single, mohammadi2019reconstruction, li2021deep}. To do so, we need a global protein interaction network~\cite{stark2006biogrid, luck2020reference}. 

\xhdr{Learning task}
On a global gene interaction network, we perform multilabel node classification to predict whether a gene is activated in a specific cell type based on single-cell RNA sequencing (scRNA-seq) experiments. In particular, if there are $N$ cell types identified in a given experiment, each gene is associated with a vector of length $N$. Given the gene interaction network and label vectors for a select number of genes, the task is to train a model that predicts every element of the vector for a new gene such that predicted values indicate the probabilities of gene activation in various cell types (Figure~\ref{fig:case-studies}a). To enable inductive learning, we split our nodes (i.e., genes) into train, validation, and test sets such that we can generalize to never-before-seen genes.

\xhdr{Impact}
Generating gene embeddings that consider differential expression at the cell type level enables predictions at a single cell resolution, with considerations for factors including disease/cell states and temporal/spatial dependencies \cite{rizvi2017single, ravindra2020disease}. Implications of such cell-type aware gene embeddings extend to cellular function prediction and identification of cell-type-specific disease features \cite{li2021deep}. For example, quantifying ligand-receptor interactions using single cell expression data has elucidated intercellular interactions in tumor microenvironments (e.g., via CellPhoneDB~\cite{efremova2020cellphonedb} or NicheNet~\cite{browaeys2020nichenet}). In fact, upon experimental validation of the predicted cell-cell interactions in distinct spatial regions of tissues and/or tumors, these studies have demonstrated the importance of spatial heterogeneity in tumors~\cite{zhang2020history}. Further, unlike most non-graph based methods, like autoencoders, GNNs are able to model dependencies (e.g., physical interactions) between proteins as well as single-cell expression~\cite{alessandri2021sparsely, tran2021fast}.

\end{regbox}

\section{Graph representation learning for genomics} \label{sec:genomics}

Diseases are classified based on the presenting symptoms of patients, which can be caused by molecular dysfunctions, such as genetic mutations. As a result, diagnosing diseases requires knowledge about alterations in the transcription of coding genes to capture genome-wide associations driving disease acquisition and progression. Graph representation learning methods allow us to analyze heterogeneous networks of multimodal data and make predictions across domains, from genomic level data (e.g., gene expression, copy number information) to clinically relevant data (e.g., pathophysiology, tissue localization).

\subsection{Leveraging gene expression measurements} \label{sec:gene_exp}

Comparing transcriptomic profiles from healthy individuals to those of patients with a specific disease informs clinicians of its causal genes. As gene expression is the direct readout of perturbation effects, changes in gene expression are often used to model disease-specific co-expression or regulatory interactions between genes. Further, injecting gene expression data into PPI networks has identified disease biomarkers, which are then used to more accurately classify diseases of interest.

Methods that rely solely on gene expression data typically transform the co-expression matrix into a more topologically meaningful form~\cite{han2019gcn, mandal2020topological, nicolau2011topology}. Gene expression data can be transformed into a colored graph that captures the shape of the data (e.g.,~using TDA \cite{nicolau2011topology}; refer to \supp{3}), which then enables downstream analysis using network science metrics and graph machine learning. Topological landscapes present in gene expression data can be vectorized and fed into a GCN to classify the disease type~\cite{mandal2020topological}. Alternatively, gene expression data can directly be used to construct disease and gene networks that are then input into a joint matrix factorization and GCN method to draw disease-gene associations, akin to a recommendation task (Section \ref{sec:gnn})~\cite{han2019gcn}. Further, applying a joint GCN, VAE, and GAN framework (Section \ref{sec:generative}) to gene correlation networks---initialized with a subset of gene expression matrices---can generate disease networks with the desired properties~\cite{yang2019conditional}.

Because gene expression data can be noisy and variational, recent advances include fusing the co-expression matrices with existing biomedical networks, such as GO annotations and PPI, and feeding the resulting graph into graph convolution layers (Section \ref{sec:gnn}) \cite{chereda2019utilizing, ma2019incorporating, rhee2018hybrid}. Doing so has enabled more interpretable disease classification models (e.g.,~weighting gene interactions based on existing biological knowledge). However, despite the utility of PPI networks, they have been reported to limit models trained solely on PPI networks because they are unable to capture all gene regulatory activities~\cite{ramirez2020classification}. To this end, graph representation learning methods, such as GNNs, have been developed to learn robust and meaningful representations of molecules despite the incomplete interactome~\cite{huang2020skipgnn} and to inductively infer new edges between pairs of nodes~\cite{liu2021indigo}.

\subsection{Injecting single cell and spatial information into molecular networks}
Single-cell RNA sequencing (scRNA-seq) data lend themselves to graph representation learning to model cellular differential processes~\cite{burkhardt2021quantifying, rizvi2017single} and disease states~\cite{ravindra2020disease}. In particular, a predominant approach to analyze scRNA-seq datasets is to transform them into gene similarity networks, such as gene co-expression networks, or cell similarity networks by correlating gene expression readouts across individual cells. Applied to such networks, graph representation learning can impute scRNA-seq data~\cite{huang2020scgnn, wang2021scgnn}, predict cell clusters~\cite{chen2021simba, wang2021scgnn}, etc. Cell similarity graphs have also been created using autoencoders by first embedding gene expression readouts and then connecting genes based on how similar their embeddings are~\cite{wang2021scgnn}. Alternatively, variational graph autoencoders produce cell embeddings and interpretable attention weights indicating what genes the model attends to when deriving an embedding for a given cell~\cite{buterez2021cellvgae}. Beyond GNNs and graph autoencoders, learning a manifold over a cell state space can quantify the effects of experimental perturbations~\cite{burkhardt2021quantifying}. 
To this end, cell similarity graphs are constructed for control and treated samples and used to estimate the likelihood of a cell population observed under a given perturbation~\cite{burkhardt2021quantifying}.

Spatial molecular profiling can measure both gene expression at the cellular level and location of cells in a tissue~\cite{marx2021method}. As a result, spatial transcriptomics data can be used to construct cell graphs~\cite{yuan2020gcng}, spatial gene expression graphs~\cite{partel2021spage2vec}, gene co-expression networks, or molecular similarity graphs~\cite{hetzel2021graph}. Creating cell neighborhood and spatial gene expression graphs require a distance metric, as edges are determined based on spatial proximity, while gene co-expression and molecular similarity graphs need a threshold applied on the gene expression data~\cite{hetzel2021graph}. From such networks, graph representation learning methods produce embeddings that capture the network topology and that can be further optimized for downstream tasks. For instance, a cell neighborhood graph and a gene pair expression matrix enable classic GNNs to predict ligand-receptor interactions~\cite{yuan2020gcng}. In fact, as ligand-receptor interactions are directed, they could be used to infer causal interactions of previously unknown ligand-receptor pairs~\cite{meinshausen2016methods,yuan2020gcng}.

\clearpage

\begin{regbox}
\normalfont\textit{\textbf{Learning representations of diseases and phenotypes (Figure~\ref{fig:case-studies}b)}}
\label{sec:genome_app} 

\xhdr{Graph dataset}
Symptoms are observable characteristics that typically result from interactions between genotypes. Physicians utilize a standardized vocabulary of symptoms, i.e., phenotypes, to describe human diseases. Thus, we model diseases as collections of associated phenotypes to diagnose patients based on their presenting symptoms. Consider a graph built from the standardized vocabulary of phenotypes, e.g.,~the Human Phenotype Ontology~\cite{robinson2008human} (HPO). The HPO forms a directed acyclic graph with nodes representing phenotypes and edges indicating hierarchical relationships between them; however, it is typically treated as an undirected graph in most implementations of GNNs. A disease described by a set of its phenotypes corresponds to a subset of nodes in the HPO, forming a subgraph of the HPO. Note that a subgraph can contain many disconnect components dispersed across the entire graph~\cite{alsentzer2020subgraph}.

\xhdr{Learning task}
Given a dataset of HPO subgraphs and disease labels for a select number of them, the task is to generate an embedding for every subgraph and use the learned subgraph embeddings to predict the disease most consistent with the set of phenotypes that the embedding represents~\cite{alsentzer2020subgraph} (Figure~\ref{fig:case-studies}b).

\xhdr{Impact}
Modeling diseases as rich graph structures, such as subgraphs, enables a more flexible representation of diseases than relying on individual nodes or edges. As a result, we can better resolve complex phenotypic relationships and improve differentiation of diseases or disorders.

\end{regbox}

\section{Graph representation learning for therapeutics} \label{sec:therapeutics}

Modern drug discovery requires elucidating a candidate drug's chemical structure, identifying its drug targets, quantifying its efficacy and toxicity, and detecting its potential side effects~\cite{barabasi2011network, guney2016network, hu2016network, rai2018network}. Because such processes are costly and time-consuming, \textit{in silico} approaches have been adopted into the drug discovery pipeline. However, cross-domain expertise is necessary to develop a drug with the optimal binding affinity and specificity to biomarkers, maximal therapy efficacy, and minimal adverse effects. As a result, it is critical to integrate chemical structure information, protein interactions, and clinically relevant data (e.g., indications and reported side effects) into predictive models for drug discovery and repurposing. Graph representation learning has been successful in characterizing drugs at the systems level without patient data to make predictions about interactions with other drugs, protein targets, side effects, diseases~\cite{cheng2018network, cheng2019genome, gaudelet2021utilizing, gysi2021network, maclean2021knowledge, zeng2022toward}.

\subsection{Modeling compound molecular graphs} \label{sec:mol_struc}

Similar to proteins, small compounds are modeled as 2D and 3D molecular graphs such that nodes are atoms and edges are bonds. Each atom and bond may include features, such as atomic mass, atomic number, and bond type, to be included in the model \cite{stokes2020deep, schnet}. Edges can also be added to indicate pairwise spatial distance between two atoms \cite{enn-s2s}, or directed with information on bond angles and rotations incorporated into the molecular graph \cite{dimenet}.

Representing molecules as graphs has improved predictions on various quantum chemistry properties. Intuitively, message passing steps (i.e.,~in GNNs (Section \ref{sec:gnn})) aggregate information from neighboring atoms and bonds to learn the local chemistry of each atom \cite{stokes2020deep}. For example, generating representations of the atoms, distances, and angles to be propagated along the molecular graph has allowed us to identify the angle and direction of interactions between atoms~\cite{dimenet}. Producing atom-centered representations based on a weighted combination of their neighbors' features (i.e., using an attention mechanism) is able to model interactions among reactants for predicting organic reaction outcomes~\cite{coley2019graph}.
Alternatively, molecular graphs have been decomposed into a ``junction tree," where each node represents a substructure in the molecule, to learn representations of both the molecular graph and the junction tree for generating new molecules with desirable properties (Section \ref{sec:generative}) \cite{jtvae}. Due to the major challenge of finding novel and diverse compounds with specific chemical properties, iteratively editing fragments of a molecular graph during training has improved predictions for high-quality candidates targeting our proteins of interest \cite{xie2021mars}.

\subsection{Quantifying drug-drug and drug-target interactions} \label{sec:drug_interact}

Corresponding to molecular structure is binding affinity and specificity to biomarkers. Such measurements are important for ensuring that a drug is effective in treating its intended disease, and does not have significant off-target effects~\cite{wieder2020compact}. However, quantifying these metrics requires labor- and cost-intensive experiments~\cite{david2020molecular, wieder2020compact}. Modeling small compounds' and protein targets' molecular structure as well as their binding affinities and specificity, for instance, using graph representation learning has enabled accelerated investigation of interactions between a given drug and protein target.

First, we learn representations of drugs and targets using graph-based methods, such as TDA \cite{alagappan2016multimodal} or shallow network embedding approaches~\cite{node2vec}. Concretely, TDA (refer to \supp{3}) transforms experimental data into a graph where nodes represent compounds and edges indicate a level of similarity between them \cite{alagappan2016multimodal}. Shallow network embedding techniques are also used to generate embeddings for drugs and targets by computing drug-drug, drug-target, and target-target similarities (Section \ref{sec:shallow}) \cite{thafar2020dtigems+}. Non-graph based methods have also been used to construct graphs that are then fed into a graph representation learning model to generate embeddings. K-nearest neighbors, for instance, is a common used method to construct drug and target similarity networks~\cite{thafar2021dti2vec}. The resulting embeddings are fed into downstream machine learning models.

Fusing compound sequence, structure, and clinical implications has significantly improved drug-drug and drug-target interaction predictions. For example, attention mechanisms have been applied on drug graphs, with chemical structures and side effects as features, to generate interpretable predictions of drug-drug interactions~\cite{ma2018drug}. Also, two separate GNNs may be used to learn representations of protein and small molecule graphs for predicting drug-target affinity~\cite{jiang2020drug}. To be flexible with other graph- and non-graph-based methods, protein structure representations generated by graph convolutions has been combined with protein sequence representations (e.g.,~shallow network embedding methods or CNNs) to predict the probability of compound-protein interactions \cite{chen2020transformercpi, quan2019graphcpi, tsubaki2019compound, lin2020deepgs}. 

\subsection{Identifying drug-disease associations and biomarkers for complex disease} \label{sec:drug_disease}

Part of the drug discovery pipeline is minimizing adverse drug events~\cite{david2020molecular, wieder2020compact}. But, in addition to high financial cost, the experiments required to measure drug-drug interactions and toxicity face a combinatorial explosion problem~\cite{david2020molecular}. Graph representation learning methods enables \textit{in silico} modeling of drug action, which allows for more efficient ranking of candidate drugs for repurposing, such as by considering gene expression data, gene ontologies, drug similarity, and other clinically relevant data regarding side effects and indications. 

Drug and disease representations have been learned on homogeneous graphs of drugs, diseases, or targets. For instance, Medical Subject Headings (MeSH) terms may be used to construct a drug-disease graph, from which latent representations of drugs and diseases are learned using various graph embedding algorithms, including DeepWalk and LINE (Section \ref{sec:shallow}) \cite{guo2020meshheading2vec}. TDA (refer to \supp{3}) has also been applied to construct graphs of drugs, targets, and diseases separately, from which representations of such entities are learned and optimized for downstream prediction~\cite{wang2019bmc}.

To emphasize the systems-level complexity of diseases, recent methods fuse multimodal data to generate heterogeneous graphs. For example, neighborhood information are aggregated from heterogeneous networks comprised of drug, target, and disease information to predict drug-target interactions (Section \ref{sec:gnn}) \cite{wan2019neodti}. In other instances, PPI networks are combined with genomic features to predict drug sensitivity using GNNs~\cite{xie2019integrating}. As a result, approaches that integrate cross-domain knowledge as a vast heterogeneous network and/or into the model's architecture seem better equipped to elucidate drug action.

\begin{regbox}
\normalfont\textit{\textbf{Learning representations of drugs and drug combinations (Figure~\ref{fig:case-studies}c)}} 
\label{sec:drug_app} 

\xhdr{Graph dataset}
Combination therapies are increasingly used to treat complex and chronic diseases. However, it is experimentally intensive and costly to evaluate whether two or more drugs interact with each other and lead to effects that go beyond additive effects of individual drugs in the combination. Graph representation learning has been equipped to leverage perturbation experiments performed across cell lines to predict the response of never-before-seen cell lines with mutation(s) of interest (e.g.,~disease-causing) to drug combinations. Consider a multimodal network of protein-protein, protein-drug, and drug-drug interactions where nodes are proteins and drugs, and edges of different types indicate physical contacts between proteins, the binding of drugs to their target proteins, and interactions between drugs (e.g.,~synergistic effects, where the effects of the combination are larger than the sum of each drug's individual effect)~\cite{zitnik2018modeling,jiang2020deep}. Such a multimodal drug-protein network is constructed for every cell line, yielding a collection of cell line specific networks (Figure~\ref{fig:case-studies}c) \cite{jiang2020deep}. 

\xhdr{Learning task}
From a single cell line's drug-protein network, we predict whether two or more drugs are interacting in the cell line \cite{jiang2020deep}. Concretely, we embed nodes of a drug-protein network into a compact embedding space such that distances between node embeddings correspond to similarities of nodes' local neighborhoods in the network. We then use the learned embeddings to decode drug-drug edges, and predict probabilities of two drugs interacting based on their embeddings. Next, we apply transfer learning to leverage the knowledge gained from one cell line specific network to accelerate the training and improve the accuracy across other cell line specific networks (Figure~\ref{fig:case-studies}c) \cite{kim2021anticancer}. Specifically, we develop a model using one cell line's drug-protein network, ``reuse'' the model on the next cell line's drug-protein network, and repeat until we have trained on drug-protein networks from all cell lines.

\xhdr{Impact}
Not only are non-graph based methods unsuited to capture topological dependencies between drugs and targets, most predictive models for drug combinations do not consider tissue or cell-line specificity of drugs. Because drugs' effects on the body are not uniform, it is crucial to account for such anatomical differences. Further, the ability to prioritize candidate drug combinations \textit{in silico} could reduce the cost of developing and testing them experimentally, thereby enabling robust evaluation of the most promising combinatorial therapies.

\end{regbox}

\section{Graph representation learning for healthcare} \label{sec:healthcare}
Graph representation learning has been used to fuse multimodal knowledge with patient records to better enable precision medicine. Two modes of patient data successfully integrated using deep graph learning are histopathological images~\cite{barisoni2020digital, chen2020pathomic} and EHRs~\cite{choi2017gram, li2020graph}.

\subsection{Leveraging networks for diagnostic imaging} \label{sec:med_img}

Medical images of patients, including histopathology slides, enable clinicians to comprehensively observe the effects of a disease on the patient's body~\cite{gurcan2009histopathological}. Medical images, such as large whole histopathology slides, are typically converted into cell spatial graphs, where nodes represent cells in the image and edges indicate that a pair of cells is adjacent in space. Deep graph learning has been shown to detect subtle signs of disease progression in the images while integrating other modalities (e.g., tissue localization~\cite{pati2020hact} and genomic features~\cite{chen2020pathomic}) to improve medical image processing.

Cell-tissue graphs generated from histopathological images are able to encode the spatial context of cells and tissues for a given patient. Cell morphology and tissue micro-architecture information can be aggregated from cell graphs to grade cancer histology images (e.g.,~using GNNs (Section \ref{sec:gnn})) \cite{adnan2020representation, anand2020histographs, chen2020pathomic, zhou2019cgc}. An example aggregation method is pooling with an attention mechanism to infer relevant patches in the image~\cite{adnan2020representation}. Further, cell morphology and interactions, tissue morphology and spatial distribution, cell-to-tissue hierarchies, and spatial distribution of cells with respect to tissues can be captured in a cell-to-tissue graph, upon which a hierarchical GNN can learn representations using these different data modalities~\cite{pati2020hact}. Because interpretability is critical for models aimed to generate patient-specific predictions, post-hoc graph pruning optimization may be performed on a cell graph generated from a histopathology image to define subgraphs explaining the original cell graph analysis \cite{jaume2020towards}.

Graph representation learning methods have also been proven successful for classifying other types of medical images.  GNNs are able to model relationships between lymph nodes to compute the spread of lymph node gross tumor volume based on radiotherapy CT images (Section \ref{sec:gnn}) \cite{chao2020lymph}. MRI images can be converted into graphs that GNNs are applied to for classifying the progression of Alzheimer's Disease \cite{an2020dynamic, song2019graph, wee2019cortical}. GNNs are also shown to leverage relational structures like similarities among chest X-rays to improve downstream tasks, such as disease diagnosis and localization~\cite{mao2022imagegcn}. Alternatively, TDA can generate graphs of whole-slide images, which include tissues from various patient sources (\supp{3}), and GNNs are then used to classify the stage of colon cancer \cite{levy2020topological}.

Further, spatial molecular profiling benefits from methodological advancements made for medical images. With spatial gene expression graphs (weighted and undirected) and corresponding histopathology images, gene expression information are aggregated to generate embeddings of genes that could then be used to investigate spatial domains (e.g., differentiate between cancer and noncancer regions in tissues)~\cite{hu2021spagcn}. Since multimodal data enables more robust predictions, GNNs (Section \ref{sec:gnn}) have been applied to generate cell spatial graphs from histopathology images and then fuse genomic and transcriptomic data for predicting treatment response and resistance, histopathology grading, and patient survival \cite{chen2020pathomic}.

\subsection{Personalizing medical knowledge networks with patient records} \label{sec:med_rec}

Electronic health records are typically represented by ICD (International Classification of Disease) codes~\cite{choi2017gram, li2020graph}. The hierarchical information inherent to ICD codes (medical ontologies) naturally lend itself to creating a rich network of medical knowledge. In addition to ICD codes, medical knowledge can take the form of other data types, including presenting symptoms, molecular data, drug interactions, and side effects. By integrating patient records into our networks, graph representation learning is well-equipped to advance precision medicine by generating predictions tailored to individual patients.

Methods that embed medical entities, including EHRs and medical ontologies, leverage the inherently hierarchical structure in the medical concepts KG~\cite{rotmensch2017learning}. Low dimensional embeddings of EHR data can be generated by separately considering medical services, doctors, and patients in shallow network embeddings (Section \ref{sec:shallow}) and graph neural networks (Section \ref{sec:gnn}) \cite{wu2019representation,mao2022medgcn}. Alternatively, attention mechanisms may be applied on EHR data and medical ontologies to capture the parent-child relationships \cite{choi2017gram,ma2018kame,sun2020disease}. Rather than assuming a certain structure in the EHRs, a Graph Convolution Transformer can even learn the hidden EHR structure~\cite{choi2020learning}.

EHRs also have underlying spatial and/or temporal dependencies~\cite{chen2020robustly} that many methods have recently taken advantage of to perform time-dependent prediction tasks. A mixed pooling multi-view self-attention autoencoder has generated patient representations for predicting either a patient's risk of developing a disease in a future visit, or the diagnostic codes of the next visit \cite{chowdhury2019mixed}. A combined LSTM and GNN model has also been used to represent patient status sequences and temporal medical event graphs, respectively, to predict future prescriptions or disease codes (Section \ref{sec:gnn}) \cite{liu2020hybrid, lee2020harmonized}. Alternatively, a patient graph may be constructed based on the similarity of patients, and patient embeddings learned by an LSTM-GNN architecture are optimized to predict patient outcomes~\cite{rocheteau2021predicting}. An ST-GCN~\cite{yan2018spatial} is designed to utilize the underlying spatial and temporal dependencies of EHR data for generating patient diagnoses~\cite{li2020graph}.

EHRs are often supplemented with other modalities, such as diseases, symptoms, molecular data, drug interactions, etc~\cite{chen2020robustly, kwak2020drug, nelson2019integrating, zhao2018emr}. A probabilistic KG of EHR data, which include medical history, drug prescriptions, and laboratory examination results, has been used to consider the semantic relations between EHR entities in a shallow network embedding method (Section \ref{sec:shallow}) \cite{li2020method}. Meta-paths may alternatively be exploited in an EHR-derived KG to leverage higher order, semantically important relations for disease classification \cite{hosseini2018heteromed}. Initializing node features for drugs and diseases using Skipgram and then applying GNN leverages multi-layer message-passing to predict adverse drug events \cite{kwak2020drug}. Moreover, combined RNNs and GNNs models have been applied to EHR data integrated with drug and disease interactions to better recommend medication combinations~\cite{shang2019gamenet}.

\begin{figure}
    \centering
    \includegraphics[width=\textwidth]{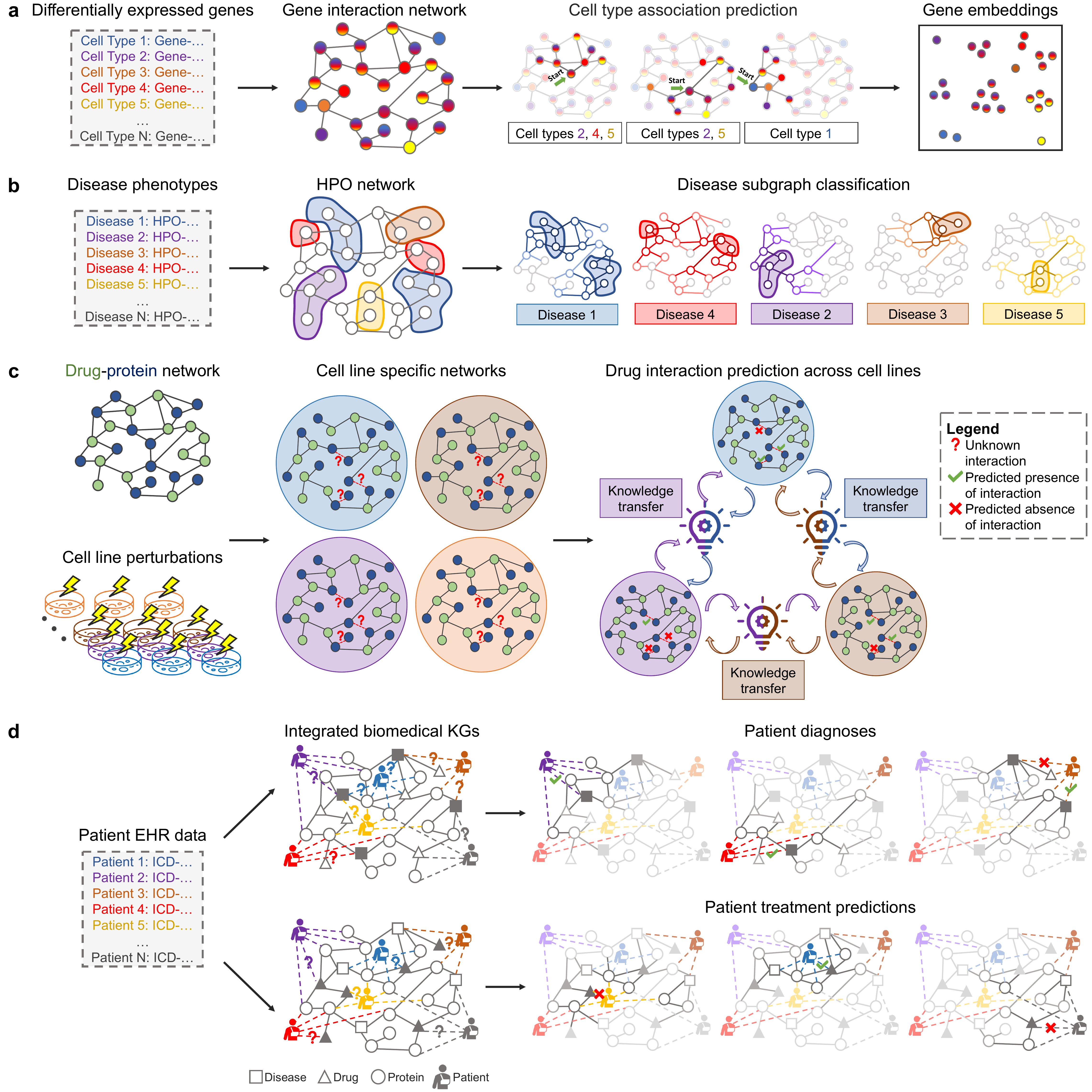}
    \caption{\textbf{Representation learning in four areas of biology and medicine.} We present a case study on \textbf{(a)} cell-type aware protein representation learning via multilabel node classification (details in Box~\ref{sec:mol_app}), \textbf{(b)} disease classification using subgraphs (details in Box~\ref{sec:genome_app}), \textbf{(c)} cell-line specific prediction of interacting drug pairs via edge regression with transfer learning across cell lines (details in Box~\ref{sec:drug_app}), and \textbf{(d)} integration of health data into knowledge graphs to predict patient diagnoses or treatments via edge regression (details in Box~\ref{sec:patient_app}).}
    \label{fig:case-studies}
\end{figure}

\begin{regbox}
\normalfont\textit{\textbf{Fusing personalized health information with knowledge graphs (Figure~\ref{fig:case-studies}d)}}
\label{sec:patient_app} 

\xhdr{Graph dataset}
To realize precision medicine, we need robust methods that can inject biomedical knowledge into patient-specific information to produce actionable and trustworthy predictions \cite{national2011toward}. Since EHRs can also be represented by networks, we are able to fuse patients' EHR networks with biomedical networks, thus enabling graph representation learning to make predictions on patient-specific features. Consider a knowledge graph, where nodes and edges represent different types of bioentities and their various relationships, respectively. Examples of relations may include ``up-/down-regulate,'' ``treats,'' ``binds,'' ``encodes,'' and ``localizes''~\cite{nelson2019integrating}. To integrate patients into the network, we create a distinct meta node to represent each patient, and add edges between the patient's meta node and its associated bioentity nodes (Figure~\ref{fig:case-studies}d).

\xhdr{Learning tasks}
We learn node embeddings for each patient while predicting (via edge regression) the probability of a patient developing a specific disease or of a drug effectively treating the patient (Figure~\ref{fig:case-studies}d) \cite{nelson2019integrating}.

\xhdr{Impact}
Precision medicine requires an understanding of patient-specific data as well as the underlying biological mechanisms of disease and healthy states. Most networks do not consider patient data, which can prevent robust predictions of patients' conditions and potential responsiveness to drugs. The ability to integrate patient data with biomedical knowledge can address such issues.

\end{regbox}

\clearpage

\section{Outlook}

As graph representation learning has aided in the mapping of genotypes to phenotypes, leveraging graph representation learning for fine-scale mapping of variants is a promising new direction \cite{wu2022network}. By re-imagining GWAS and expression Quantitative Trait Loci (eQTL) studies \cite{umans2020disease} as networks, we can already begin to discover biologically meaningful modules to highlight key genes involved in the underlying mechanisms of a disease \cite{wang2020disease}. We can alternatively seed network propagation with QTL candidate genes~\cite{wu2022network}. Additionally, because graphs can model long-range dependencies or interactions, we can model chromatin elements and the effects of their binding to regions across the genome as a network~\cite{dekker2015long, lanchantin2020graph}. We could even reconstruct 3D chromosomal structures by predicting 3D coordinates for nodes derived from a Hi-C contact map~\cite{hovenga2021hic}. Further, as spatial molecular profiling has enabled profound discoveries for diseases, graph representation learning repertoire to analyze such datasets will continue to expand. For instance, with dynamic GNNs, we may be able to better capture changes in expression levels observed in single cell RNA sequencing data over time or as a result of a perturbation \cite{burkhardt2021quantifying, ding2022temporal}.

Effective integration of healthcare data with knowledge about molecular, genomic, disease-level, and drug-level data can help generate more accurate and interpretable predictions about the biological systems of health and disease~\cite{fortelny2020knowledge}. Given the utility of graphs in both the biological and medical domains, there has been a major push to generate knowledge graphs that synthesize and model multi-scaled, multi-modal data, from genotype-phenotype associations to population-scale epidemiological dynamics.  In public health, spatial and temporal networks can model space- and time-dependent observations (e.g.,~disease states, susceptibility to infection \cite{machens2013infectious}) to spot trends, detect anomalies, and interpret temporal dynamics.

As artificial intelligence tools implementing graph representation learning algorithms are increasingly employed in clinical applications, it is essential to ensure that representations are explainable~\cite{gnnexplainer}, fair~\cite{agarwal2021towards}, and robust~\cite{zhang2020gnnguard}, and that existing tools are revisited in light of algorithmic bias and health disparities~\cite{obermeyer2019dissecting}.

\section*{Acknowledgements}

We gratefully acknowledge the support of NSF under Nos.~IIS-2030459 and IIS-2033384, US Air Force Contract No.~FA8702-15-D-0001, Harvard Data Science Initiative, Amazon Research Award, Bayer Early Excellence in Science Award, AstraZeneca Research, and Roche Alliance with Distinguished Scientists Award. M.M.L.~is supported by T32HG002295 from the National Human Genome Research Institute and a National Science Foundation Graduate Research Fellowship. Any opinions, findings, conclusions or recommendations expressed in this material are those of the authors and do not necessarily reflect the views of the funders. The authors declare that there are no conflicts of interests.

\bibliographystyle{abbrv}
\bibliography{reference}





\end{document}


\maketitle
\thispagestyle{empty}
\spacing{1.3}

\begin{center}
{\Large Supplementary Information for\\[3mm]
{\bf Graph Representation Learning in Biomedicine}\\[3mm]}
Michelle M. Li$^{1,3}$, Kexin Huang$^{2}$, and Marinka Zitnik$^{3,4,5,\ddag}$ \\[1mm]  
\normalsize{$^{1}$Bioinformatics and Integrative Genomics, Harvard Medical School, Boston, MA 02115, USA} \\  
\normalsize{$^{2}$Health Data Science, Harvard T.H. Chan School of Public Health, Boston, MA 02115, USA} \\
\normalsize{$^{3}$Department of Biomedical Informatics, Harvard Medical School, Boston, MA 02115, USA} \\
\normalsize{$^{4}$Broad Institute of MIT and Harvard, Cambridge, MA 02142, USA} \\
\normalsize{$^{5}$Harvard Data Science Initiative, Cambridge, MA 02138, USA} \\[4mm]  
\normalsize{$\ddag$Corresponding author. Email: marinka@hms.harvard.edu}
\end{center}

\vspace{1cm}

{\spacing{1}
\noindent This PDF file includes:
\begin{description}[labelsep=2em, align=left, itemsep=0em]
\item[] Supplementary Notes 1 to 4
\item[] Supplementary Figure 1
\item[] Supplementary References
\end{description}
\textbf{}
\vspace{1cm}

}

\clearpage
\setcounter{page}{2} 
\tableofcontents

\listoffigures  
\clearpage

\clearpage

\section{Further Information on Graph Notation and Definitions}\label{sec:def}

\xhdr{Graph-theoretic elements}
Graphs consist of the following key elements: 
\begin{itemize}[leftmargin=*]
    \item \textit{Node} $v$ represents a biomedical entity, ranging from atoms to patients. 
    
    \item \textit{Edge} $e_{u,v}$ is a relation or link between node entities $u$ and $v$, such as a bond between atoms, an affinity between molecules, a disease association between phenotypes, and a referral between a patient and a doctor. We denote the edge set in the graph as $\mathcal{E}$ and the complementary non-edge set as $\mathcal{E}^c$. The edge can be directed, oriented such that it points from a source (or head node) to a destination (or tail node). The edge can also be undirected, where the nodes have two-way relations. 
    
    \item \textit{Graph} $G = (\mathcal{V}, \mathcal{E})$ consists of a collection of nodes $\mathcal{V}$ that are connected by an edge set $\mathcal{E}$, such as a molecular graph or protein interaction network. Adjacency matrix $\mathbf{A}$ is commonly used to represent a graph, where each entry $\mathbf{A}_{u,v}$ is 1 if nodes $u, v$ are connected, and 0 otherwise. $\mathbf{A}_{u,v}$ can also be the edge weight between nodes $u,v$. We denote the number of the nodes $\vert \mathcal{V} \vert = n$ and the number of edges $\vert \mathcal{E} \vert = m$. 
    
    \item \textit{Subgraph} $S = (\mathcal{V}_S, \mathcal{E}_S)$ is a subset of a graph $G = (\mathcal{V}, \mathcal{E})$, where $\mathcal{V}_S \subseteq \mathcal{V}, \mathcal{E}_S \subseteq \mathcal{E}$. Examples include disease modules in a protein interaction network or communities in a patient-doctor referral network.
    
    \item \textit{Node Feature} $\mathbf{x}_v \in \mathbb{R}^d$ describes  attributes of node $v$. The node feature matrix is denoted as $\mathbf{X} \in \mathbb{R}^{n \times d}$. Similarly, we can have edge features $\mathbf{x}_{u,v}^e \in \mathbb{R}^c$ for edge $e_{u,v}$ collected together into edge feature matrix $\mathbf{X}^e \in \mathbb{R}^{m \times c}$.
    
    \item \textit{Label} is a target value associated with either a node $Y_v$, an edge $Y_e$, a subgraph $Y_S$, or a graph $Y_G$.
\end{itemize}

\xhdr{Walks and paths in graphs} A walk of length $l$ from node $v_1$ to node $v_l$ is a sequence of nodes and edges $v_1 \xrightarrow{e_{1,2}} v_2 \cdots v_{l-1} \xrightarrow{e_{l-1,l}} v_l$. A (simple) path is a type of walks where all nodes in the walk are distinct. For every two nodes $u, v$ in the graph $G$, we define the distance $d(u, v)$ as the length of the shortest path between them. In a heterogeneous graph, a meta-path is a sequence of node types $V_i$ and their edge types $R_{i,j}$: $V_1 \xrightarrow{R_{1,2}} V_2 \cdots V_{l-1} \xrightarrow{R_{l-1,l}} V_l$. 

\xhdr{Local neighborhoods} For a node $v$, we denote its neighborhood $\mathcal{N}(v)$ as the collection of nodes that are connected to $v$, and its degree is the size of $\mathcal{N}(v)$. The $k$ hop neighborhood of node $v$ is the set of nodes that are exactly $k$ hops away from node $v$: $\mathcal{N}^k(v) = \{u | d(u, v) = k\}$. 

\xhdr{Graph types}
The following types of graphs are commonly considered to model complex biomedical systems.

\begin{itemize}[leftmargin=*]

\item \textit{Simple, weighted, and attributed graphs.} A simple graph $G = (\mathcal{V}, \mathcal{E})$ is fully described by nodes $\mathcal{V}$ and edges $\mathcal{E}$. For example, a set of binary PPIs gives rise to an unweighted PPI network. Further, nodes and edges can be accompanied by single-dimensional (e.g., edge weights) or multi-dimensional attribute vectors describing node and edge properties. For example, each node in a cell-cell interaction network can have a gene expression attribute vector encoding the gene's expression profile.

\item \textit{Multimodal or heterogeneous graphs.} Multimodal or heterogeneous graphs consist of nodes of different types (node type set $\mathcal{A}$) connected by diverse kinds of edges (edge type set $\mathcal{R}$). For example, in a drug-target-disease interaction network, nodes represent $\mathcal{A} = \{\textrm{drugs}, \textrm{proteins}, \textrm{diseases}\}$ and edges indicate $\mathcal{R} = \{ \textrm{drug-target binding}, \textrm{disease-associated mutations}, \textrm{treatments} \}$. 

\item \textit{Knowledge graphs.} A biomedical knowledge graph is a heterogeneous graph that captures knowledge retrieved from literature and biorepositories. A knowledge graph is given by a set of triplets $(u, r, v) \in \mathcal{V} \times \mathcal{R} \times \mathcal{V}$, where nodes $u,v$ belong to node types $\mathcal{A}$ and are connected by edges with type $r \in \mathcal{R}$.

\item \textit{Multi-layer graphs.} Multi-layer graphs capture hierarchical relations by grouping individual networks into different layers. Formally, we have a set of networks $G_1, \cdots, G_n$ and $l$ layers where each layer corresponds to a set of networks. Different layers can represent distinct contexts, such as tissues or diseases. Edges can also be added across layers. For example, each tissue can be represented with a tissue-specific PPI network, and PPIs for every tissue can be organized by tissue taxonomy, where each layer corresponds to a tissue taxonomy. Inter-layer edges can be connected for the same proteins across tissues~\cite{zitnik2017predicting}. 

\item \textit{Temporal graphs.} Biomedical systems evolve over time. A temporal graph consists of a sequence of graphs $G_1, \cdots, G_T$ ordered by time, where at each time step $t$, we observe a subset of all nodes and their activity. For example, a human brain can be modeled as a temporal graph of brain regions showing task-related increases in neural activity at time $t$ (e.g., greater activity during an experimental task than during a baseline state) and linked based on functional connectivity at $t$.

\item \textit{Spatial graphs.} Nodes or edges in a spatial graph are spatial elements usually associated with coordinates in one, two, or three dimensions, e.g., a spatial representation of cell-cell interactions in the 3-dimensional (3D) Euclidean tissue environment or a 3D point cloud of the protein's atomic coordinates. Spatial graphs are defined by having nodes or edges with spatial locations. This small modification to aspatial graphs has profound effects on how these graphs are used and interpreted because a spatial graph is a location map of points with the constraints of space rather than an abstract structure.

\end{itemize}
The graph types described above can be combined to give rise to new objects, such as multi-layer spatial graphs or multimodal temporal graphs.

\section{Overview of Graph Machine Learning Tasks}\label{sec:other_ml}

We divide machine learning tasks on biomedical graphs into three broad categories: graph prediction, latent graph learning, and graph generation. Each category is associated with several individual graph machine learning tasks.

\xhdr{Canonical graph prediction} Graph prediction aims to predict a label in the graph. The label can be associated with any unit of the graph. There are four canonical graph learning tasks: 
(1) \textit{Node classification/regression} aims to find a function $f: \mathcal{V} \rightarrow Y_\mathcal{V}$ that predicts the label of a node in the graph; 
(2) \textit{Link prediction} aims to find a function $f: \mathcal{E} \cup \mathcal{E}^c  \rightarrow \{0, 1\}$) to predict whether there exists a link between a given pair of nodes in the graph; 
(3) \textit{Edge classification/regression} aims to find a function $f: \mathcal{E} \rightarrow Y_\mathcal{E}$ that predicts the label of an edge; 
(4) \textit{Graph classification/regression} aims to find a function  $f: \mathcal{G} \rightarrow Y_\mathcal{G}$ that maps each graph in a graph set to the correct label.

\xhdr{Other graph prediction tasks} In addition to the four standard prediction tasks on graphs, there are additional tasks that are particularly important for biomedical graphs: 
(1) \textit{Module detection} aims to detect a subgraph module in the graph that contributes to a variable; 
(2) \textit{Clustering or community detection} aims to partition the graphs into a set of subgraphs such that each subgraph contains similar nodes; 
(3) \textit{Subgraph classification/regression} aims to predict a label for the subgraph or module; 
(4) \textit{Dynamic graph prediction} aims to perform the above prediction tasks in a sequence of dynamic graphs.

\xhdr{Latent graph learning} While graph prediction tasks predict given the graph-structured data, latent graph learning aims to obtain a function  $f: \mathcal{V}, \mathbf{X} \rightarrow \mathcal{E}$ to learn the underlying graph structure (e.g., edges) given only the nodes and their feature attributes. The learned graph can be used to (1) perform graph prediction tasks; (2) obtain the inherent topology of the data; and (3) generate latent low-dimensional representations of the feature attributes.

\xhdr{Graph generation} The objective of graph generation is to generate a never-before-seen graph $G$ with some properties of interest. Given a set of training graphs $\mathcal{G}$ with certain shared characteristics, the task is to learn a function $f: \mathcal{G} \rightarrow \mathcal{D}_\mathcal{G}$ to obtain a distribution $\mathcal{D}_\mathcal{G}$ that characterizes the training graphs. Then, the learned distribution can be used to generate a new graph $G'$, which has the same characteristics as or optimized properties compared to the training graphs.

\section{Further Details on Representation Learning Approaches} \label{sec:other_rep}

We here review four important graph representation learning approaches beyond those surveyed in the main text. In particular, we describe:
\begin{itemize}
\item \textit{Graph theoretic techniques (Section~\ref{sec:heuristics}):} Networks model relations among real world subjects. Such relations form patterns of structures in the network. These patterns can be quantified by expert-defined statistics to characterize the role of nodes, links, or subgraphs. These statistics can be used to represent elements in the graph.
\item \textit{Network propagation methods (Section~\ref{sec:diffusion}):} Nodes in a graph influence each other along the paths. Diffusion measures these spreads of influences. By aggregating diffusion from neighboring nodes to the target node, a diffusion profile captures the local connectivity patterns of the target node. 
\item \textit{Topological data analysis (Section~\ref{sec:tda}):} A dataset has an underlying structure. Topological data analysis (TDA) analyzes the topology of the data to generate an underlying graph structure. Two main frameworks exist. One is called persistent homology~~\cite{edelsbrunner2008persistent}, which obtains a vector that quantifies various topological shapes at different spatial resolutions. The other is called Mapper~\cite{nicolau2011topology,singh2007topological}, which first clusters topologically similar data into a node, where similarity is defined by a filter function, and then connects the clusters as the backbone of the data topology. The output is a graph that characterizes the topology of data.
\item \textit{Manifold learning (Section~\ref{sec:manifold}):} Real-world data is usually high-dimensional. To better interpret them, a mapping to find the low-dimensional characterization of the data is ideal. This mapping is called manifold learning, or non-linear dimensionality reduction.
\end{itemize}

\subsection{Graph theoretic techniques} \label{sec:heuristics} 

Networks model relations among real world subjects. Such relations form patterns of structures in the network. Among the network science community, many have studied these patterns of graph structures, and proposed graph statistics to measure their characteristics. We summarize such statistics into the following three categories: node-, link-, and subgraph-level statistics. See Figure~\ref{fig:SI_method}a for an illustration of network statistics.

\xhdr{Node-level techniques} The goal of node-level statistics is to measure the role of a node $u$ in a graph. For instance, betweenness~\cite{freeman1977set} calculates the number of shortest paths that pass through the node. A node with high betweenness is called a bottleneck because it controls the information flow of the network. Various centrality statistics are proposed to measure the various roles of a node regarding its structure and function. For example, k-core~\cite{kitsak2010identification} measures the position of the node in the graph.

\xhdr{Link-level techniques} Statistics have been demonstrated as a powerful tool for link prediction. They are used to represent the likelihood that a link exists between two nodes. Classic statistics include common neighbor index, Adamic–Adar index, resource allocation index, etc \cite{gao2015link}. Most of them rely on the homophily principle. However, \cite{kovacs2019network} recently showed that the protein-protein interaction (PPI) network does not follow homophily because interacting proteins are not necessarily similar, and similar proteins do not necessarily interact. They propose L3, which calculates the number of paths of length 3, and it shows strong performance in learning PPI networks. For a complete list of link-level graph statistics methods, we refer readers to~\cite{liu2020computational}. 

\xhdr{Subgraph-level techniques} Many subnetwork patterns recur in the network. Such patterns are called motifs, and they are shown to be the basic blocks of complex networks. For instance, a feed-forward loop is an important three-node motif in gene regulatory networks~\cite{milo2002network}. They have functional roles, such as increasing the response to signals~\cite{martin2016drivers}. Individualized motifs for each network can also be computed through frequent subgraph mining algorithms~\cite{jiang2013survey}.

\subsection{Network diffusion} \label{sec:diffusion} 

Nodes in a graph influence each other along the paths. Diffusion measures these spreads of influences. The typical resulting outcomes of interest include a scalar $\psi$ between every node $v$ to the source node $u$ that measures the influence, or a diffusion profile $\psi_u$ for source node $u$, which captures the local connectivity patterns (Figure~\ref{fig:SI_method}b). Many have studied the effect of diffusion in physics, economics, epidemiology and various formulations have been proposed~\cite{cowan2004network,raj2012network,akbarpour2018diffusion}. On a very related line of work, label propagation leverages connected links to propagate labels~\cite{wang2007label}.

\xhdr{Diffusion state distance} One effective method for biomedical networks is the diffusion state distance~\cite{cao2013going}, which first calculates the number of times a random walk starting at source node will visit a destination node given a fixed number of steps, and iterates this process for every destination node in the graph.

\xhdr{Unsupervised extension} Recent efforts, such as GraphWave~\cite{GraphWave}, adopt an unsupervised learning method to learn an embedding for each node by leveraging heat wavelet diffusion patterns. The resulting embeddings allow nodes residing in different parts of a graph to have similar structural roles within their local network topology.

\subsection{Topological data analysis} \label{sec:tda} 

\begin{figure}
    \centering
    \includegraphics[width=0.85\textwidth]{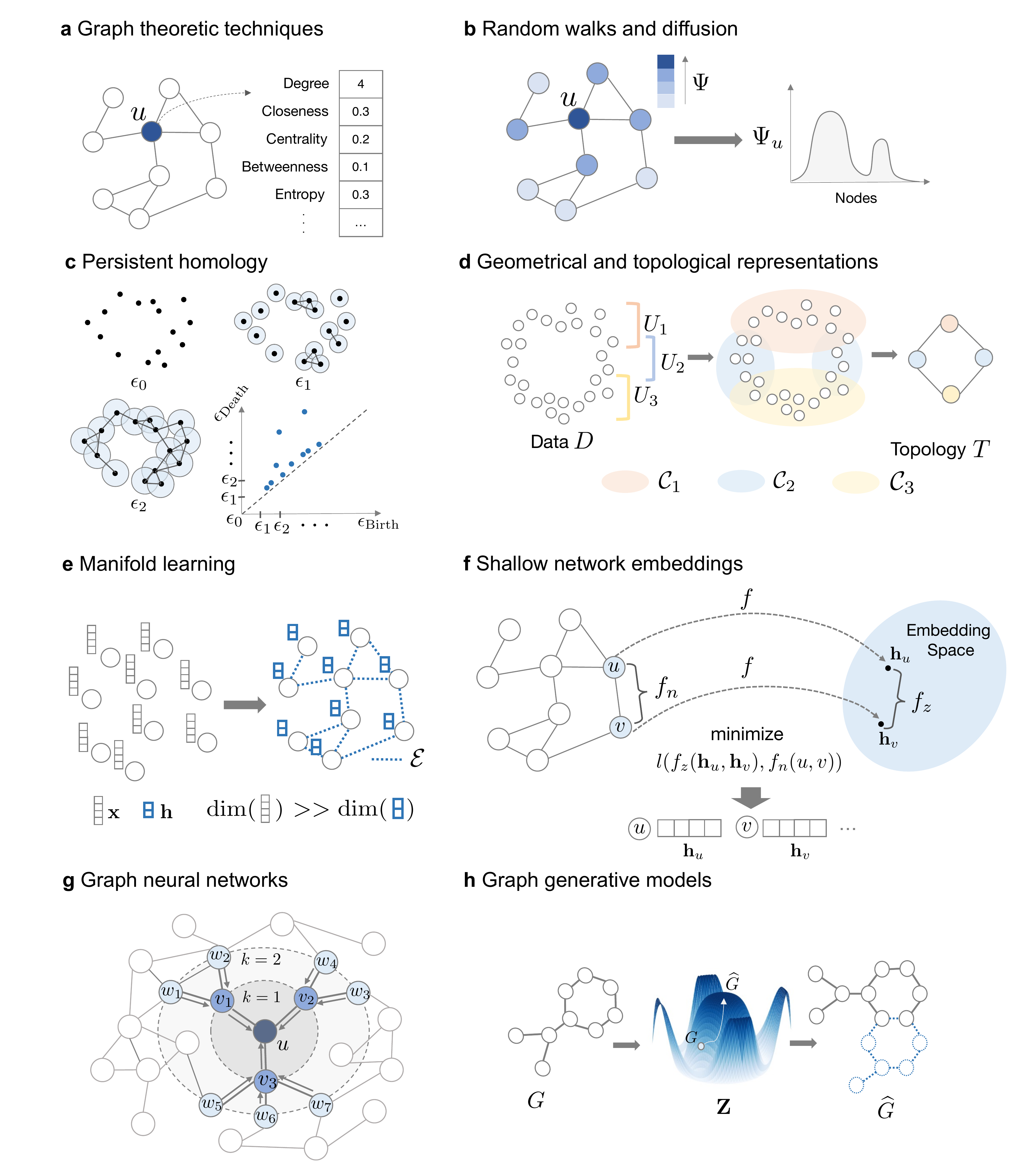}
    \caption[\textbf{Predominant graph learning paradigms.}]{\textbf{Predominant graph learning paradigms.} Graph machine learning is a large field with a diverse set of methods. Here, we summarize and categorize seven major graph machine learning methods paradigms. \textbf{(a)} Graph theoretic techniques compute a deterministic value that describes patterns in the graph; \textbf{(b)} diffusion process captures the importance and influence of nodes through network diffusion; \textbf{(c-d)} topological data analysis provides summarized views of the shape of the data; \textbf{(e)} manifold learning aims to obtain the underlying graph structure of data and a low-dimensional embedding; \textbf{(f)} shallow network embeddings generate node representations though direct encoding of node similarities in the input graph; \textbf{(g)} graph neural networks learn graph embeddings through supervised signals by neural networks; \textbf{(h)} generative models generate novel graphs that have desirable properties. Note that panels (c) and (d) are two major diagrams of TDA.}
    \label{fig:SI_method}
\end{figure}

For a large and high-dimensional dataset $D$, it is hard to directly gauge their characteristics or obtain a summary of the data. However, all data have an underlying shape or topology $T$, which can be considered as a network. Topological data analysis (TDA) analyzes the topology of the data to generate an underlying graph structure. There are two major diagrams in TDA: persistent homology (Figure~\ref{fig:SI_method}c) and mapper (Figure~\ref{fig:SI_method}d).

\xhdr{Persistent homology} Persistent homology~\cite{edelsbrunner2008persistent} obtains a vector that quantifies various topological shapes at different spatial resolutions. As the resolution scale expands, the noise and artifacts would disappear while the important structure persists. Recent works have used neural networks on top of persistent diagrams to learn augmented topological features~\cite{hofer2017deep,carriere2020perslay}. \cite{hofer2020graph} propose a differentiable persistent homology layer in the network to make any GNN topology-aware. 

\xhdr{Mapper} While persistent homology provides a vector of topology, mapper generates a topology graph~\cite{nicolau2011topology,singh2007topological}. This graph is obtained by first clustering topologically similar data into a node, where similarity is defined by a filter function, and then connecting the clusters as the backbone of the data topology. The resulting shape can be used to visualize the data and understand data subtypes~\cite{de2015head} and trajectories of development~\cite{madhobi2019visual}. Mapper is highly dependent on the filter function, and it is usually constructed with domain expertise. Recent works have integrated neural networks to automatically learn the filter function from the data~\cite{deepgraphmapper}.

\subsection{Manifold learning} \label{sec:manifold}  

Real-world data is usually high-dimensional. To better interpret them, a mapping to find the low-dimensional characterization of the data is ideal. This mapping is called manifold learning, or non-linear dimensionality reduction (Figure~\ref{fig:SI_method}e). Note that the underlying manifold can be considered as a weighted network such that higher weights are assigned to edges between data points that are closer in the manifold. 

In a typical setting, we only have a set of data points, or nodes $u_1, \cdots, u_n$, and their associated attributes $\mathbf{x}_1, \cdots, \mathbf{x}_n$ without any connections among them. To learn the underlying manifold using graphs, the first step is to construct an edge set $\mathcal{E}$ that connects nodes given some distance measure. With this connectivity graph, one approach is to apply a graph statistics operator to directly compute the low-dimensional embedding, such as laplacian eigenmap~\cite{belkin2001laplacian} and isomap~\cite{tenenbaum2000global}. Another approach is to optimize various kinds of cost functions to generate low-dimensional embeddings that preserve distance measures on the graph, such as t-SNE~\cite{maaten2008visualizing}. However, such methods are all multi-stage processes in which the outcome depends on the defined distance measures. Recently, a line of research has emerged that can generate embeddings in an end-to-end learnable manner, where the manifold is learned by the signals from the downstream prediction task~\cite{halcrow2020grale,wang2019dynamic}.

\section{Further Details on Mathematical Formulations} \label{sec:mat-formulations}

\vspace{5mm}

\begin{ourbox} \textit{\textbf{Graph theoretic techniques}} 

Graph theoretic techniques are functions that map network components to real-values representing aspects of graph structure, such as node proximity~\cite{menche2015uncovering,baryshnikova2016systematic,agrawal2018large} and node centrality~\cite{goh2007human}. We use betweenness as an example.

\xhdr{Example} For node $u$ in graph $G$, the betweenness is calculated as: $B_u = \sum_{s \neq u \neq t} \frac{\sigma_{s,t}(u)}{\sigma_{s,t}}$, where $\sigma_{s,t}$ is the number of shortest paths between nodes $s$ and $t$, and $\sigma_{s,t}(u)$ is the number of shortest paths that pass through node $u$. Basically, the larger the betweenness, the larger the influence of this node $u$ on the network. 

\end{ourbox}

\begin{ourbox} \textit{\textbf{Network diffusion}}

Network diffusion computes the network influence signatures based on propagation on the networks. We use Diffusion State Distance (DSD)~\cite{cao2013going} as an example. 

\xhdr{Example} For a node $u$, we calculate its diffusion distance $D(u,v_i)$ from every node $v_i \in \mathcal{V}$ as the expected number of times that $p_u$, a random walk of length $k$ starting from $u$, will visit $v_i$. Formally, $D(u,v_i) = \mathbb{E}[sign(v_i, p_u)]$, where $sign(v_i, p_u)$ is $1$ if node $v_i \in p_u$ and $0$ otherwise. So, we obtain a vector $D(u) = (D(u,v_1), \cdots , D(u,v_n))$. Then, the DSD between nodes $u$ and $v$ is defined as $DSD(u, v) = \Vert D(u) - D(v)\Vert_1$, the L1-norm of the diffusion vector difference. Intuitively, DSD measures the differences in the node influence from every other node to see if they have similar local connectivity. 
\end{ourbox}

\begin{ourbox} \textit{\textbf{Geometric representations along preassigned guiding functions called filters}}

A mapper generates the data topology $T$ from the data $D$ \cite{singh2007topological,wang2019bmc}. A standard mapper procedure consists of the following steps: 

\begin{enumerate}
    \item \xhdr{Reference Map} Given a set of data $D$, we first define a continuous filter function $f: D \rightarrow Z$ that assigns every data point in $D$ to a value in $Z$.
    \item \xhdr{Construction of a Covering} A finite covering $\mathbb{U} = \{U_\alpha\}_{\alpha \in A}$ is constructed on $Z$. Each cover consists of a set of points $D_\alpha$, and is the pre-image of the cover $D_\alpha = f^{(-1)}(U_\alpha)$.
    \item \xhdr{Clustering} For each subset $D_\alpha$, we apply a clustering algorithm $C$ that generates $N_\alpha$ clusters.
    \item \xhdr{Topology Graph} Each cluster forms a node. If two clusters share data points in $D$, then an edge is formed. The resulting graph is the topology graph $T$ of data $D$. Formally, the topology graph is defined as the nerve of the cover by the path-connected components. 
\end{enumerate}

\end{ourbox}

\begin{ourbox} \textit{\textbf{Shallow network embedding}}

Shallow network embedding generates a mapping that preserves the similarity in the network \cite{hosseini2018heteromed}. Formally, typical shallow network embeddings are learned in the following three steps:

\begin{enumerate}
    \item \textbf{Mapping to an embedding space.} Given a pair of nodes $u, v$ in network $G$, we obtain a function $f$ to map these nodes to an embedding space to generate $\mathbf{h}_u$ and $\mathbf{h}_v$.
    \item \textbf{Defining network similarity.} We next define the network similarity as $f_n(u,v)$, and the embedding similarity as $f_z(\mathbf{h}_u,\mathbf{h}_v)$.
    \item \textbf{Computing loss.} Then, we define the loss $\mathcal{L}(f_n(u,v), f_z(\mathbf{h}_u,\mathbf{h}_v))$, which measures whether the embedding preserves the distance in the original networks. Finally, we apply an optimization procedure to minimize the loss $\mathcal{L}(f_n(u,v), f_z(\mathbf{h}_u,\mathbf{h}_v))$. 
\end{enumerate}

\end{ourbox}

\begin{ourbox} \textit{\textbf{Persistent homology}}

Given a dataset $D$, persistent homology generates a persistence diagram (often referred to as a barcode) that captures the significant topological features in $D$, such as connected components, holes, and cavities~\cite{hofer2019learning, bubenik2015statistical}. It consists of the following steps:

\begin{enumerate}
    \item \xhdr{Construction of the Rips Complex} Consider each data point in the original dataset $D$ as a vertex. For each pair of vertices $u, v$, create an edge if the distance between them is at most $\epsilon$, i.e., $\mathcal{E} = \{(u, v) | d(u, v) \le \epsilon\}$, given the distance metric $d$. Consider a monotonically increasing sequence of $\epsilon_0, \cdots, \epsilon_n$, we then generate a filtration of Rips complexes $G_0, \cdots, G_n$.
    \item \xhdr{Homology} Homology characterizes topological structures (e.g., 0-th order homology measures connected components, 1-st order homology measures holes, 2-nd order homology measures voids). A class in a $k$-th order homology is an instantiation of the homology. In each Rips complex, we can track the various homology classes, which capture various topological structures. For a rigorous definition of homology, we refer reader to~\cite{edelsbrunner2010computational}.
    \item \xhdr{Persistence Diagrams} At each $\epsilon$, we denote the emergence of a new homology class $a$ as its birth using the current $\epsilon$. Similarly, we denote the disappearance of a previous homology class $a$ as its death. Thus, for each homology class $a$, we can represent them as $(\epsilon_{Birth}, \epsilon_{Death})$. After $\epsilon$ reaches the end of the sequence, we have a set of 2-dimensional points $(x, y)$, each corresponding to the birth and death of a homology class. The persistence diagram is a plot of these points. The farther away from the diagonal, the longer the lifespan of the homology class, implying that the structure is an important shape of the data topology. 
\end{enumerate}

\end{ourbox}

\begin{ourbox} \textit{\textbf{Manifold learning}}

The goal of manifold learning is to learn a low-dimensional embedding that captures the data manifold from high-dimensional data \cite{burkhardt2019quantifying}. In the following, we use isomap~\cite{tenenbaum2000global} as an example. 

\xhdr{Example} First, given a set of data points with high-dimensional feature vectors $\mathbf{x}_1, \cdots, \mathbf{x}_n$, we apply a $k$-nearest neighbor algorithm and connect the nearest neighbors to form a neighborhood graph $G$. Next, we calculate the distance matrix $\mathbf{D}$, where each entry $\mathbf{d}_{i,j}$ is the length of the shortest path between nodes $i$ and $j$ in the neighborhood graph. Then, we apply an optimization algorithm to obtain a set of low-dimensional vectors $\mathbf{h}_1, \cdots, \mathbf{h}_n$ that minimizes $\sum_{i < j} \left( \Vert \mathbf{h}_i - \mathbf{h}_j \Vert - d_{i,j}\right)^2 $. 

\end{ourbox}

\begin{ourbox} \textit{\textbf{Graph neural networks}}

GNNs learn compact representations or embeddings that capture network structure and node features \cite{han2019gcn, protein_design, xie2019integrating}. A GNN generates outputs through a series of propagation layers~\cite{enn-s2s}, where propagation at layer $l$ consists of the following three steps: 

\begin{enumerate}
    \item \textbf{Neural message passing.} The GNN computes a message $\mathbf{m}^{(l)}_{u,v} = \textsc{Msg}(\mathbf{h}_u^{(l-1)}, \mathbf{h}_v^{(l-1)})$ for every linked nodes $u,v$ based on their embeddings from the previous layer $\mathbf{h}_u^{(l-1)}$ and $\mathbf{h}_v^{(l-1)}$.
    \item \textbf{Neighborhood aggregation.} The messages between node $u$ and its neighbors $\mathcal{N}_u$ are aggregated as $\hat{\mathbf{m}}^{(l)}_{u} = \textsc{Agg}({\mathbf{m}^{(l)}_{uv} | v \in \mathcal{N}_u})$.
    \item \textbf{Update.} The GNN applies a non-linear function to update node embeddings as $\mathbf{h}^{(l)}_u = \textsc{Upd}(\hat{\mathbf{m}}^{(l)}_{u}, \mathbf{h}^{(l-1)}_u)$ using the aggregated message and the embedding from the previous layer.
\end{enumerate}

\end{ourbox}

\begin{ourbox} \textit{\textbf{Generative modeling}} 

Generative models optimize and learn data distributions in order to generate graphs with desirable properties. In the following, we use VGAE~\cite{VGAE} as an example. 

\xhdr{Example} VGAE is a graph extension of variational autoencoders~\cite{kingma2013auto}. Given a graph $G$ with adjacency matrix $\mathbf{A}$ and node features $\mathbf{X}$, we use two GNNs to encode the graph and generate the latent mean vector $\mu_\mathbf{Z}^u$ and the log-variance $\mathrm{log}(\sigma_\mathbf{Z}^u)$ parameters for each node $u$. Formally, $\mu_\mathbf{Z} = \mathrm{GNN}_\mu(\mathbf{A}, \mathbf{X}) \in \mathbb{R}^{\vert \mathcal{V} \vert \times d}$, and $\mathrm{log}(\sigma_\mathbf{Z}) = \mathrm{GNN}_\sigma(\mathbf{A}, \mathbf{X}) \in \mathbb{R}^{\vert\mathcal{V}\vert \times d}$. We can then obtain the latent distribution $q(\mathbf{Z} | G) \sim \mathcal{N} (\mu_\mathbf{Z}, (\mathrm{log}(\sigma_\mathbf{Z})))$, where we can draw a latent embedding sample $\mathbf{Z} \in \mathbb{R}^{\vert \mathcal{V} \vert} \times d$ from the latent distribution. Then, given the latent embedding, we feed into a probabilistic decoder $p_\theta(\hat{\mathbf{A}}|\mathbf{Z})$ to generate a network $\hat{G}$ with adjacency matrix $\hat{\mathbf{A}}$. In VGAE, the decoder is a dot product, i.e., $p_\theta(\hat{\mathbf{A}}_{uv} = 1|\mathbf{Z}) = \mathrm{sigmoid}(\mathbf{z}_u^T\mathbf{z}_v)$. This way, we obtain a newly generated graph $\hat{G}$. Finally, we optimize the weight using the variational reconstruction loss $\mathcal{L} = \mathbb{E}_{q(\mathbf{Z}|G)}[p(G|\mathbf{Z})] - \mathrm{KL}(q(\mathbf{Z}|G)\Vert p(\mathbf{Z}))$.

\end{ourbox}

\clearpage



\clearpage
\section*{Supplementary References}
\label{sec:refs}
\addcontentsline{toc}{section}{\nameref{sec:refs}}
\bibliographystyle{naturemag}
\bibliography{supplementary}